\RequirePackage{fix-cm}
\documentclass{article}

\usepackage{arxiv}

\usepackage[utf8]{inputenc} 
\usepackage[T1]{fontenc}    
\usepackage{hyperref}       
\usepackage{url}            
\usepackage{booktabs}       
\usepackage{amsfonts}       
\usepackage{nicefrac}       
\usepackage{microtype}      
\usepackage{lipsum}		
\usepackage{graphicx}
\usepackage{natbib}
\usepackage{doi,subcaption}
\usepackage{amsmath,amsthm}
\usepackage{amssymb,dsfont}
\usepackage{xcolor,mathtools}

\newtheorem{rem}{\emph{Remark}}
\newtheorem{defn}{Definition}
\newtheorem{thm}{Theorem}

\newtheorem{lemma}{Lemma}
\newtheorem{assumptions}{Assumptions}
\allowdisplaybreaks

\begin{document}
\makeatletter
\def\blfootnote{\gdef\@thefnmark{}\@footnotetext}
\makeatother

\title{The Role of Mutual Information in Variational Classifiers 
}


\author{
Matias Vera  \\
Facultad de Ingeniería, Universidad de Buenos Aires\\
CSC, CONICET\\
Buenos Aires, Argentina \\
  \texttt{mvera@fi.uba.ar} \\
   \And
Leonardo Rey Vega \\
Facultad de Ingeniería, Universidad de Buenos Aires\\
 CSC, CONICET\\
Buenos Aires, Argentina \\
  \texttt{lrey@fi.uba.ar} \\
  \And
Pablo Piantanida \\
International Laboratory on Learning Systems (ILLS)\\
McGill - ETS - Mila - CNRS - Universit\'e Paris Saclay - CentraleSupelec\\
Montréal, Canada\\
  \texttt{pablo.piantanida@centralesupelec.fr} \\
}  



\date{}

\maketitle
\begin{abstract}
{\bf This paper was accepted for publication to Machine Learning (Springer).}\\
Overfitting data is a well-known phenomenon related with the generation of a model that mimics too closely (or exactly) a particular instance of data, and may therefore fail to predict future observations reliably. In practice, this behaviour is controlled by various--sometimes based on heuristics--regularization techniques, which are motivated by upper bounds to the generalization error. In this work, we study the generalization error of classifiers relying on stochastic encodings which are trained on the cross-entropy loss, which is often used in deep learning for classification problems. We derive bounds to the generalization error showing that there exists a regime where the generalization error is bounded by the mutual information between input features and the corresponding representations in the latent space, which are  randomly generated according to the encoding distribution. Our bounds provide an information-theoretic understanding of generalization in the so-called class of variational classifiers, which are regularized by a Kullback-Leibler (KL) divergence term. These results give theoretical grounds for the highly popular KL term in variational inference methods that was already recognized to act effectively as a regularization penalty. We further observe connections with well studied notions such as Variational Autoencoders, Information Dropout, Information Bottleneck and Boltzmann Machines. Finally, we perform numerical experiments on MNIST, CIFAR and other datasets and show that mutual information is indeed highly representative of the behaviour of the generalization error.



\end{abstract}

\section{Introduction}
\label{sec:intro}

The major challenge in representation learning is to learn the different explanatory factors in a given dataset. Learning models are often guided by the objective of optimizing performance on training data when the real objective is to generalize well to unseen data. Generalization error, i.e. the difference between expected and empirical risk, is the standard measure for quantifying the capacity of an algorithm to generalize learned patterns from seen data to unseen ones \citep[Chapter 2]{MohriRostamizadehTalwalkar18}. An upper bound that dominates the aforementioned error can often provide an indication of what is a good regularization term for empirical risk minimization. In this way, investigating the information-theoretic impact of different regularization techniques on the generalization error is could help in the understanding of these methods. In this work, we investigate upper bounds to the generalization error in terms of the mutual information between input features and latent representations intended to classifiers. These classifiers rely on stochastic encodings trained on the cross-entropy loss. More precisely, we present a PAC-learning result that analyzes the impact of regularization techniques based on the information shared between inputs $X$ and the corresponding latent representations $U_{(X)}$. We observe interesting connections with Variational Autoencoders (VAEs) \citep{kingma2013auto} and the so-called Information Bottleneck method (IB) \citep{Tishby1999information}, among others.

\subsection{Our Contributions}

{The main result of the paper can be loosely summarized (for the specific details see Theorem \ref{thm:regularizar}) as follows:
\begin{equation}\label{eq:teorema_basicaversion}
   \big|  \mbox{true cross-entropy} - \mbox{empirical cross-entropy} \big | \leq \mathcal{O}\left( \sqrt{\mathcal{I}\big( U_{(X)} ; X \big)  }  \;\frac{\log (n)}{\sqrt{n}}\right)
\end{equation}
with high probability and where $n$ is the sample size of the training set and $\mathcal{I}\big( U_{(X)} ; X \big) $ indicates the mutual information between the random input features $X$ and the latent representations $U_{(X)}$ using a stochastic encoding $q_{U|X}: X \mapsto U_{(X)}$.} This result motivates formally the use of the above mutual information as a regularizing term to control the {amount} of information conveyed by latent representations about input features, leading to  the minimization of the following objective: 
\begin{equation}
\mbox{empirical cross-entropy} + \lambda\cdot \mathcal{I}\big( U_{(X)} ; X \big), \ \ \forall\,\lambda\geq 0, 
\label{eq-loss-VC}
\end{equation}
which was commonly implemented via an appropriate (empirical) upper-bound to the mutual information using Kullback Leibler (KL) divergence: 
\begin{equation}
\mathcal{I}\big( U_{(X)} ; X \big)\leq \mathbb{E}_{p_X} \left[\text{KL}( q_{U|X}(\cdot|X) \| q_U )\right], \ \ \forall\,q_U,
\label{eq-loss-VC-bound}
\end{equation}
and where $q_U$ is some known prior (often normal) distribution on the latent representation space. Although a theoretical understanding of the connection between the above mutual information and the generalization of the cross-entropy loss remains elusive in the literature, the impact of the multiplier $\lambda$ together with the KL bound \eqref{eq-loss-VC-bound} in the training objective \eqref{eq-loss-VC} has been shown empirically to improve performance of some deep learning algorithms by \citep{8253482, kingma2013auto}, among others works.  


Our analysis of the generalization error is framed within the classification problem in a specific setting. There are two main ingredients that characterize our model: (a) we focus on randomized encodings, which allows us not only to study deterministic algorithms, but also to cover graphical methods such as variational classifiers \citep{maggipinto2020betavariational} and Restricted Boltzmann Machines (RBMs) \citep{Hinton_guia}; and (b) our analysis assumes cross-entropy as the loss to be trained. It is worth mentioning some of our motivation behind these assumptions. {From a purely mathematical view-point, by allowing stochastic encodings rules, 
we extend our searching space \citep{Yamanishi92} given that deterministic encoders/decoders are a particular case of stochastic ones}. On the other hand, the cross-entropy loss is often used to train state-of-art deep learning algorithms for classification problems. Furthermore, the cross-entropy is even employed as a performance metric for deep learning algorithms ending with a soft-max layer \citep[Section 6.2.2]{Goodfellow-et-al-2016}, which provides a notion of how likely each class is (i.e., the so called calibration). Unfortunately, modern neural networks may be poorly calibrated even when they are accurate \citep{guo17}. Unlike error probability, cross-entropy at testing time is an excellent metric that evaluates both accuracy and calibration.

Theorem~\ref{thm:regularizar} gives the rigorous statement of the PAC-learning bound on the generalization error which is indeed the basis of this work. Our bound depends on several ingredients: a mutual information term between the inputs  and the latent representations generated from them and a measure of the decoder efficiency, among others items, and it is inspired by the seminal work of \citep{Xu2012}. Although our results may not lead to the tightest bounds, neither in terms of statistical scaling with respect to the sample size nor of the involved constants, these attempt to reflect the importance of information-theoretic concepts in representation learning and the different trade-offs that can be established between information measures and quantities of interest in statistical learning. 

{An empirical investigation of the interplay between the generalization error and the above described mutual information is mainly performed (but not limited to) on datasets of natural images.} These simulations show the ability of the mutual information in \eqref{eq-loss-VC-bound} to predict the behavior of generalization for the case of three rather simple but well-known representation models: (a) the gaussian variational classifier \citep{maggipinto2020betavariational} which is a classification version of standard variational autoencoder \citep{kingma2013auto}; (b) the log-normal encoder presented in the Information Dropout scheme \citep{2016arXiv161101353A}; and (c)  encodings based on simple RBMs \citep{Hinton_guia}.  These numerical results indicate  that the mutual information between inputs features and their corresponding representations is clearly correlated to the behaviour of the generalization error properties of some learning algorithms, which raises the need for further studies in this respect. 

\subsection{Related Literature}

The idea of relating the generalization capacity of moderns machine learning algorithms to information measures is not new. In the last decade, the topic has received considerable attention \citep{Vincent2010Stacked, Russo_Zou_2015, kingma2013auto, 2016arXiv161101353A, Halbersberg20}. Perhaps, one of the most exciting approach is the so-called Information Bottleneck (IB) method by \citep{DBLP:journals/corr/TishbyZ15}. The IB principle postulates that the mutual information between inputs and the corresponding latent representations is related to the overfitting problem,  establishing a trade-off between accuracy and a measure of compression which is interpreted as an information complexity measure. Although the information-theoretic principles are well-grounded, the experimental validation of this trade-off still remains a challenging task because of the underlying difficulties in estimating mutual information with statistical confidence in high-dimensional spaces, as it was recently reported by \citep{pichler2020estimation}. Hopefully, in some cases closed-form approximations for the mutual information can be derived as it is the case for VAEs and RBMs due to the presence of stochastic encodings of latent representations. 

Statistical rates on the empirical estimates corresponding to IB trade-offs have been reported by~\citep{Shamir:2010:LGI:1808343.1808503} and a deviation bound related to the cross-entropy loss was reported by \citep{ourisit18}. In \citep{DBLP:journals/corr/Shwartz-ZivT17}, it is empirically shown that deep networks undergo two phases consisting of an initial fitting phase and a subsequent compression phase of the features where the last phase is causally related to the well-known generalization performance of deep neural networks. However, subsequent works by~\citep{DBLP:journals/corr/abs-1802-09766,saxe2018} report that none of these claims hold true in general. In other words, regularization based on \eqref{eq-loss-VC} could improve performance but there is no enough evidence to conclude that compression phase is responsible for well-known generalization capabilities of deep neural networks.

The robust framework defined by \citep{Xu2012} provides a novel approach, different from complexity or stability \citep{Bousquet_Elisseeff_2002} arguments, for studying the performance of learning algorithms in terms of the generalization error. It is showed that feed-forward neural networks are robust provided that the $L1$-norm of the weights in each layer is bounded. In this context,  \citet{Sokolic2017RobustLM} explore the Jacobian matrix of the model as a bound to the generalization error, extending results to convolutional networks as well. \citet{pmlr-v54-sokolic17a} showed that the bound of the generalization error of a stable invariant classifier is much smaller than the one of a robust non-invariant classifier.

From a different perspective, generalization in deep neural networks was also studied by \citep{Neyshabur2017GeometryOO}, where it is shown that some other forms of capacity control--different  from  network  size--plays  a central  role  in  learning these networks. The work of \citep{DBLP:journals/corr/ZhangBHRV16} concluded  that the effective capacity of several successful neural network architectures is large enough to shatter the training data. Consequently, these models are in principle rich enough to memorize the full training data.


{
The rest of the paper is organized as follows. In Section \ref{sec:learning}, we introduce the underlying learning model following \citep{2016arXiv161101353A}. In Section \ref{sec:otros}, we explore some of the existent connections with others approaches  presented in the literature.  In Section  \ref{sec:mainbound}, we present our  main result. Sections \ref{sec:casogaussiano} and \ref{sec:particion} discuss some results in view of the assumptions made in the main theorem and further discuss the results. Finally, in Section \ref{sec:experimentos} we provide numerical evidence for some selected models and concluding remarks are relegated to Section \ref{sec:conclusion}. Major mathematical details is given in Appendices \ref{sec:proof_main_theorem} to  \ref{app:auxiliary-results}, and complementary simulations are presented in Appendix \ref{app:experimentos}.
}

\subsubsection*{Notation and Conventions}

Table \ref{tab:notation} presents the most relevant symbols used across this paper.
\begin{table}
	\caption{Table of symbols}
	\label{tab:notation}       
	\begin{tabular}{ll}
		\hline\noalign{\smallskip}
		Symbol & Meaning\\
		\noalign{\smallskip}\hline\noalign{\smallskip}
		$\mathcal{X}$, $\mathcal{Y}$ and $\mathcal{U}$ & Input, target and representation spaces.\\
		$P_Y$ & Pmf for the labels $Y$.\\
		$p_X$ and $p_{XY}$ & Pdf for the inputs $X$ and joint pdf for $X$ and $Y$.\\
		$P_X$ and $P_{X|Y}(\cdot|y)$ & Probability measure associated to $p_X$ and $p_{X|Y}(\cdot|y)$.\\
		$\mathbb{E}[\cdot]$ and $\text{Var}(\cdot)$& Mathematical expectation and variance.\\
		$\theta\in\Theta$ & Parameter and parameter space.\\
		$f_\theta\coloneqq(q^\theta_{U|X}$, $Q^\theta_{\widehat{Y}|U})$ & Randomized encoder and decoder.\\
		$q_U^\theta$, $Q^\theta_{Y|U}$ & pdfs and pmfs generated from $p_{XY}q_{U|X}^\theta$.\\
		$\ell_\theta(x,y)$ & Cross-entropy loss function generated by ($q^\theta_{U|X}$, $Q^\theta_{\widehat{Y}|U}$).\\
		$\tilde{\ell}_\theta(x,y)$ & Cross-entropy loss function generated by ($q^\theta_{U|X}$, $Q^\theta_{Y|U}$).\\
		$\mathcal{S}_n$ & Training dataset.\\
		$\mathcal{L}\left(f_\theta \right)$ & Expected risk.\\
		$\mathcal{L}_\text{emp}\left(f_\theta,\mathcal{S}_n\right)$& Empirical risk.\\
		$\widehat{\theta}_n$ & Parameter than minimize the empirical risk.\\
		$\mathcal{E}_{\text{gen-err}}(\mathcal{S}_n)$& Generalization error.\\ 
		$\mathcal{O}(a_n)$, $o(a_n)$ & big-o and small-o notation.\\
		$|\cdot|$ & Cardinality.\\
		$d_x$, $d_u$ & Input and representation dimensions.\\
		$\Omega=\left\{\ell_\theta(x,y):\;\theta\in\Theta\right\}$& Loss function class.\\
		$\textrm{KL}(\cdot\|\cdot)$& Kullback Leibler divergence.\\
		$\mathcal{I}(\cdot;\cdot)$, $\mathcal{H}(\cdot)$, $\mathcal{H}_d(\cdot)$ & Mutual information, entropy and differential entropy.\\
		$\mathcal{F}_{\Omega}^{(\varepsilon)}$& Finite covering parameter set.\\
		$\{\mathcal{K}_k^{(y)}\}_{k=1}^K,\{x^{(k,y)}\}_{k=1}^K$&Cells and centroids of input-space discretization for each $y\in\mathcal{Y}$.\\
		$\epsilon^\theta(K)$, $r(K)$ & Basic elements of the input-space discretization.\\
		$P^D_{XY}$, $P_X^D$, $P_{X|Y}^D$ & pmfs generated from the input-space discretization.\\
		$q_U^{D,\theta}$, $Q_{Y|U}^{D,\theta}$, $q^{D,\theta}_{U|Y}$, $Q^{D,\theta}_{X|U}$& pdfs and pmfs generated from $P_{XY}^Dq_{U|X}^\theta$.\\
		$T^\theta(x,y)$&Difference between the loss function and an artificial loss function.\\
		$\mathcal{E}_{\text{gap}}(f_\theta,\mathcal{S}_n)$&Error-gap.\\
		$\mathcal{L}^D(f_\theta)$&Discretized version of expected risk.\\ $\mathcal{L}_{\textrm{emp}}^D(f_\theta,\mathcal{S}_n)$&Discretized version of empirical risk.\\ $\mathcal{E}_{\textrm{gap}}^D(f_\theta,\mathcal{S}_n)$& Discretized version of error-gap.\\
		$\widehat{P}^D_{XY}$, $\widehat{P}_X^D$, $\widehat{P}_{X|Y}^D$, $\widehat{P}_Y$& Empirical estimation over $\mathcal{S}_n$ (ocurrence rate).\\
		$\widehat{q}_U^{D,\theta}$, $\widehat{Q}_{Y|U}^{D,\theta}$, $\widehat{q}^{D,\theta}_{U|Y}$, $\widehat{Q}^{D,\theta}_{X|U}$& pdfs and pmfs generated from $\widehat{P}_{XY}^Dq_{U|X}^\theta$.\\
		$\ell^D_\theta(x,y)$& Cross-entropy loss function generated by $(q_{U|X}^\theta,Q_{Y|U}^{D,\theta})$.\\
		$\|\cdot\|_2$ & Norm-2 for finite vectors.\\
		$\langle\cdot,\cdot\rangle$ & Euclidean inner product.\\
		\noalign{\smallskip}\hline
	\end{tabular}
\end{table}

{
\section{Definitions}
In this section we present the base learning model that we will study in this paper and its relationship with other woks presented in the literature.
}
\subsection{The Learning Model}
\label{sec:learning}

We are interested in the problem of pattern classification consisting in the prediction of the unknown class matching an observation. This framework is defined by three main elements: (1) the source model that defines the data probability law; (2) the representation space of hidden units which allows us to divide the decision rule into an encoder and a decoder; and (3) the learning model architecture, which determines the set of possible solutions.
{
\begin{enumerate}
	\item \textbf{Source model:} Let $(\mathcal{Y}, \mathcal{B}_Y)$ and $(\mathcal{X}, \mathcal{B}_X)$ be two measurable spaces, where $\mathcal{Y}$ is discrete with
	$|\mathcal{Y} |<\infty $, $\mathcal{X} \subseteq \mathbb{R}^{d_x}$ and $\mathcal{B}_X$, $\mathcal{B}_Y$ are appropriate $\sigma$-algebras. Let $P_{X|Y}(\cdot | y)$, $y\in\mathcal{Y}$, be a collection of probability distributions such that for every  $y\in\mathcal{Y}$, $P_{X|Y}(\cdot| y)$ is a probability measure on $(\mathcal{X} , \mathcal{B}_X)$, and for every fixed $\mathcal{B} \in \mathcal{B}_X$, $P_{X|Y}(\mathcal{B} | \cdot )$  is measurable on  $(\mathcal{Y}, \mathcal{B}_Y)$ with respect to  $P_{Y}$. $P_Y$ and $P_{X|Y}(\cdot| y)$ induce the probability measure $P_{XY}$ on $(\mathcal{X}\times \mathcal{Y}, \mathcal{B}_X\times \mathcal{B}_Y)$, and where for every $y\in\mathcal{Y}$ we assume that there exists the corresponding probability density function denoted by $p_{X|Y}(x|y)$ with respect to the usual Lebesgue measure in $\mathcal{X}$\footnote{This assumption is not strictly needed and it is only assumed for simplicity.}.  Using a slight abuse the notation we will use $p_{XY}(x,y)$ for $p_{X|Y}(x|y)P_{Y}(y)$.
	\item \textbf{Representation model:} Let $(\mathcal{U}, \mathcal{B}_U)$ be a measurable space with  representation space $\mathcal{U}\subseteq\mathbb{R}^{d_u}$, and let $Q_{U|X}(u | x)$, $x\in\mathcal{X}$, be a collection of probability distributions such that for every  $x\in\mathcal{X}$, $Q_{U|X}(\cdot | x)$ is a probability measure on $(\mathcal{U} , \mathcal{B}_U)$, and for every fixed $\mathcal{B} \in \mathcal{B}_U$, $Q_{U|X}(\mathcal{B} | \cdot )$  is measurable on  $(\mathcal{X}, \mathcal{B}_X)$ with respect to  $P_{X}$. Finally, let $Q_{\widehat{Y}|U}(y |u)$, $u\in\mathcal{U}$, be a collection of probability distributions 
	 such that for every  $u\in\mathcal{U}$, $Q_{\widehat{Y}|U}(\cdot | u)$ is a probability measure on $(\mathcal{Y} , \mathcal{B}_Y)$, and for every fixed $\mathcal{B} \in \mathcal{B}_Y$, $Q_{\widehat{Y}|U}(\mathcal{B} | \cdot )$  is measurable on  $(\mathcal{U}, \mathcal{B}_U)$ with respect to  $P_{U}$. Typically, distribution $Q_{U|X}(\cdot | x)$ has the appropriate regularity conditions in order to have a density $q_{U|X}(\cdot|x)$ with respect to the Lebesgue measure in $\mathcal{U}$ for almost every $x\in\mathcal{X}$. The above definitions induce a stochastic decision rule $Q_{\widehat{Y}|X}(y|x)$ that is given by the marginalization over $\mathcal{U}$ of $q_{U|X}(\cdot|x)Q_{\widehat{Y}|U}(y|\cdot)$. 
	\item \textbf{Class of encoding distributions and predictors:}  The encoder/decoder selection is not arbitrary, there is a parametric model $\mathcal{H} \coloneqq  
	\big \{f_\theta:\, \theta\in\Theta\subset  \mathbb{R}^{l} \big\}$ with $l\in\mathbb{N}$ where $f_\theta \equiv\big(q_{U|X}^{\theta},Q_{\widehat{Y}|U}^{\theta}\big)$. Sometimes, when we will need to refer to the random representations generated by the encoder $q_{U|X}^{\theta}$ we will use $U_{\theta}$. When we will need to focus on specific parameter details for the the encoder and decoder separately, we will write them $\theta\equiv \left(\theta_E,\theta_D\right)$ with $\theta_E$ and $\theta_D$ the encoder and decoder parameters respectively and $f_\theta \equiv\left(q_{U|X}^{\theta_E},Q_{\widehat{Y}|U}^{\theta_D}\right)$.
\end{enumerate}
}
We will concern ourselves with learning representation models (randomized encoders) and inference models (randomized  decoders) from randomly generated samples.  The problem of finding a good classifier can be  divided into that of simultaneously finding an encoder $q^\theta_{U|X}$  that maps raw data to a (latent) space $\mathcal{U}$ and a soft-decoder $Q^\theta_{\hat{Y}|U}$ which maps the representation to a probability distribution on labels $\mathcal{Y}$. These  mappings induce a classifier: 
\begin{equation}
Q^\theta_{\hat{Y}|X}(y|x) = \mathbb{E}_{q^\theta_{U|X}}\left[Q^\theta_{\hat{Y}|U} (y|U)|X=x\right],  \label{eq-clasifier}
\end{equation}
\begin{rem}
	In the standard methodology with deep representations, we consider $L$ randomized encoders ($L$ layers) $\{q^\theta_{U_l|U_{l-1}}\}_{l=1}^L$ with $U_0\equiv X$. Although this appears at first to be more general, it can be casted formally using the one-layer case formulation  induced by the marginal distribution that relates  the input and the final $L$-th output layer. Therefore, results on the one-layer formulation also apply to the $L$-th layer formulation and thus, the focus of the mathematical developments will be on the one-layer case.  
\end{rem}
{This representation contains several cases of interest, as feed-forward neural networks as well as genuinely graphical model cases (e.g., VAE or RBM). The computation of \eqref{eq-clasifier} requires marginalizing over $u \in\mathcal{U}$ which is in general computationally prohibitive in practice. When $\mathcal{U}$ is a discrete space this marginalization involves sums over a large number of terms and when $\mathcal{U}$ lives in a high-dimensional real space the marginalization involves numerically expensive integration if no closed form expressions are available. A variational upper bound is used to rewrite $-\log Q^\theta_{\hat{Y}|X}(y|x)$ in the following form:}
\begin{equation}\label{eq:ourloss}
-\log Q^\theta_{\hat{Y}|X}(y|x)\leq   \mathbb{E}_{q^\theta_{U|X}} \left[ -\log Q^\theta_{\hat{Y}|U}(y|U)|X=x\right], 
\end{equation}
which simply follows by applying Jensen inequality~\citep{cover}. The above equation suggest using the  \emph{cross-entropy} as a loss-function:
\begin{equation}\label{eq:loss}
\ell_\theta(x,y) \coloneqq  \mathbb{E}_{q_{U|X}^\theta}\left[\left.-\log Q_{\widehat{Y}|U}^\theta(y|U)\right|X=x\right].
\end{equation}
Equality in \eqref{eq:ourloss} holds for the feed-forward neural network case, where $U=g(X)$ \emph{almost surely} for some $g:\mathcal{X}\rightarrow\mathcal{U}$, and $Q^\theta_{\hat{Y}|X}(y|x)=Q^\theta_{\hat{Y}|U}(y|g(x))$, {i.e., the cross-entropy in \eqref{eq:loss} includes the usual loss function used for feed-forward models.} 

The learner's goal is  to select $f_\theta\coloneqq(q^\theta_{U|X}$, $Q^\theta_{\widehat{Y}|U})$ minimizing the so-called expected risk: 
	\begin{equation}
\mathcal{L}\left(f_\theta\right)\coloneqq  \mathbb{E}_{p_{XY}}\left[\ell_\theta(X,Y)\right].
\end{equation}
Since $p_{XY}$ is unknown, the risk cannot be directly computed and it is usual to consider the empirical risk. Let the training set $\mathcal{S}_n=\{(X_k,Y_k)\}_{k=1}^n$, the empirical risk is defined by
	\begin{equation}
\mathcal{L}_\text{emp}\left(f_\theta,\mathcal{S}_n\right)\coloneqq  \frac{1}{n}\sum_{(x,y)\in\mathcal{S}_n}\ell_\theta(x,y), 
\label{eq:empirical_risk}
\end{equation}
where the parameter selection $\widehat{\theta}_n\coloneqq h(\mathcal{S}_n)$ is chosen to  minimize \eqref{eq:empirical_risk}:
	\begin{equation}\label{eq:thetahatn}
\widehat{\theta}_n\coloneqq \mathop{\arg\min}_{\theta\in\Theta}\;\mathcal{L}_\text{emp}\left(f_\theta,\mathcal{S}_n\right). 
\end{equation}
The generalization error is defined by
{
\begin{equation}\label{eq:generror}
\mathcal{E}_{\text{gen-err}}(\mathcal{S}_n)\coloneqq \mathcal{L}\big(f_{\widehat{\theta}_n} \big)-\mathcal{L}_\text{emp}\big(f_{\widehat{\theta}_n},\mathcal{S}_n\big).
\end{equation} 
Note that $\mathcal{L}\big(f_{\widehat{\theta}_n} \big)=\mathcal{L}_\text{emp}\big(f_{\widehat{\theta}_n},\mathcal{S}_n\big)+\mathcal{E}_{\text{gen-err}}(\mathcal{S}_n)$, i.e. the generalization error can be thought as a
measure of the overfitting introduced by empirical risk minimization.
}

\subsection{Connections with Related Results}
\label{sec:otros}

There is an interesting connection between the risk minimization of the cross-entropy loss and the IB principle presented by \citep{Tishby1999information}: 

\begin{defn}[Information Bottleneck]\label{def-IB-algo}
	The IB method~\citep{Tishby1999information} consists in finding $q_{U|X}$ that minimizes  the functional: 
	\begin{equation}
	\hspace{-0.2cm}\mathcal{L}_{\text{IB}}^{(\lambda)}(q^{\theta}_{U|X})\coloneqq  \mathcal{H}(Y|U^\theta)+\lambda\cdot \mathcal{I}\big( U^\theta_{(X)};X \big) ,  \label{eq-IB-function}
	\end{equation}
	for a suitable multiplier $\lambda\geq 0$, where
	\begin{align}
	q^\theta_U(u)&\coloneqq\mathbb{E}_{p_X}\left[q^\theta_{U|X}(u|X)\right],\label{eq-:qu_cont}\\ {Q}^\theta_{Y|U}(y|u)&\coloneqq\frac{\mathbb{E}_{p_X}\left[q^\theta_{U|X}(u|X)P_{Y|X}(y|X)\right]}{q^\theta_U(u)}. \label{eq-empirical-decoder}
	\end{align}
\end{defn}
{In a parametric classification problem, the IB can be interpreted as a minimization of the conditional entropy $\mathcal{H}( Y|U^\theta )$ with a regularization term $\mathcal{I}\big( U_{(X)}^\theta ; X \big)$. Clearly, $\mathcal{H}( Y|U^\theta )$ is independent of the parametric decoder $Q^\theta_{\hat{Y}|U}$ and it is a lower bound of the risk:
\begin{align}
{\mathcal{L}}(q^\theta_{U|X},Q^\theta_{\hat{Y}|U})&=\mathbb{E}_{q^\theta_U} \left[ \textrm{KL}\left(Q^\theta_{Y|U}\|Q^\theta_{\hat{Y}|U} \right)\right]+\mathcal{H}( Y|U^\theta )\label{eq:efectoib}\\
&\geq \mathcal{H}( Y|U^\theta ) \label{eq:L=D+H}\\
&=\mathcal{L}\left(q^\theta_{U|X},Q^\theta_{Y|U}\right),
\end{align}
where the equality in \eqref{eq:L=D+H} holds if and only if $Q^\theta_{\hat{Y}|U}=Q^\theta_{Y|U}$ \emph{almost surely}, i.e. in order to minimize the risk, the learner should choose the decoder induced by the encoder in \eqref{eq-empirical-decoder} while trying to minimize the resulting risk with respect to the encoder.} Typically, in real-world applications, the probability measure $p_{XY}$ is unknown, and usually the learning algorithm chooses a decoder belonging to an appropriately defined parametric class which not necessarily contain  \eqref{eq-empirical-decoder}. Either way, the decoder induced by the encoder $Q^\theta_{Y|U}$ will always exist and with it the cross-entropy loss induced only by the encoder is defined by:
	\begin{equation}
\tilde{\ell}_\theta(x,y) \coloneqq  \mathbb{E}_{q_{U|X}^\theta}\left[\left.-\log Q_{{Y}|U}^\theta(y|U)\right|X=x\right].
\label{eq:optimal_decoder}
\end{equation}
Additionally, we observe that using an arbitrary $\tilde{q}_U\in\mathcal{P}(\mathcal{U})$ (a so called prior) in a classical variational setting,
\begin{align}
\mathcal{L}_{\textrm{IB}}^{(\lambda)}(q^\theta_{U|X})&=\mathcal{H}(Y |U^\theta )+\lambda\cdot \left[ \mathbb{E}_{{p}_{X}} \left[ \textrm{KL}\big(q^\theta_{U|X} \| \tilde{q}_U \big)\right] -\textrm{KL}\big(q^\theta_{U} \| \tilde{q}_U\big)\right]\\
&\leq \mathcal{H}(Y |U^\theta ) +\lambda\cdot  \mathbb{E}_{{p}_{X}}\left[\textrm{KL}\big(q^\theta_{U|X} \| \tilde{q}_U \big)\right]\\
& \equiv   \mathcal{L}^{(\lambda)}_{\textrm{VA}}(q^\theta_{U|X},\tilde{q}_U). \label{eq-expression-VA}
\end{align}
The surrogate risk in \eqref{eq-expression-VA} is closely related with a slightly more general form of VAE discussed in~\citep{kingma2013auto} and \citep{2016arXiv161101353A},  where the latent space is regularized using a normal prior $\tilde{q}_U$. {In particular, the approach where an hyperparameter $\lambda$ is included in the cost function is known as $\beta$-VAE \citep{higgins17}.  
\begin{rem}
The loss function \eqref{eq:loss} not only matches with the classic loss\\ $-\log Q^\theta_{\hat{Y}|X}(y|x)$ typically used for feed-forward networks, but can also be interpreted as the supervised version of the $\beta$-VAE loss function:
\begin{equation}
\mathbb{E}_{q_{U|X}^\theta}\left[\left.-\log q_{\widehat{X}|U}^\theta(x|U)\right|X=x\right]+\beta\cdot\textrm{KL}\big(q^\theta_{U|X}(\cdot|x) \| \tilde{q}_U \big) ,
\label{eq:loss_beta_VAE}
\end{equation}
where $q_{U|X}^\theta$ and $q_{\widehat{X}|U}^\theta$ are the encoder and decoder respectively. If the Kullback Leibler term is interpreted as a regularization one, the main term is a surrogate of the basic loss function:
\begin{equation}
\mathbb{E}_{q_{U|X}^\theta}\left[\left.-\log q_{\widehat{X}|U}^\theta(x|U)\right|X=x\right]\geq -\log q^\theta_{\hat{X}|X}(x|x),
\label{eq:loss_beta_VAE2}
\end{equation}
where $q^\theta_{\hat{X}|X}(a|b)=\mathbb{E}_{q_{U|X}^\theta}\left[\left.q_{\widehat{X}|U}^\theta(a|U)\right|X=b\right]$. In this sense, \eqref{eq:loss}, is the supervised analogue to the first term in (\ref{eq:loss_beta_VAE}) where $Q^\theta_{\hat{Y}|U}(y|U)$ is replaced by $q_{\widehat{X}|U}^\theta(x|U)$. Similarly, (\ref{eq:loss_beta_VAE2}) is the analogue to (\ref{eq:ourloss}) with $Q^\theta_{\hat{Y}|X}(y|x)$ replaced by $q^\theta_{\hat{X}|X}(x|x)$. 
\end{rem}
Another connection to the present work can be found in \citep{Xu2012}, in which a $K-$elements partition $\mathcal{K}$ of the $\mathcal{X}$ space is proposed for obtaining, among others, the following result:
\begin{align}
\mathbb{P}\left(\mathcal{E}_{\textrm{gen-err}}({\mathcal{S}_n})\leq\inf_{\mathcal{K}}d_\theta(\mathcal{K})+M_\theta\sqrt{\frac{2K\log(2)+2\log(1/\delta)}{n}}\right)\geq1-\delta\label{eq:xubound}
\end{align}
with probability at least $1-\delta$, where $\mathcal{S}_{n}$ denotes the $n$ size training set, $M_\theta$ is the maximum value of the loss function, $K$ is the size of the partition and $d_\theta(\mathcal{K})$ is the maximum diameter of any element in the partition measured in terms of the cost function. $(K,d_\theta(\mathcal{K}))$ are parameters that define the robustness of the decision rule. Analogous to these two parameters, our bound, presented in the next section introduces two magnitudes $(\epsilon(\mathcal{K}), r(\mathcal{K}))$ that can be linked to some aspects of the robustness of the learning problem (see Section \ref{sec:mainbound}).}

Further works where mutual information plays a significant role in generalization are reported within the well-known framework of Bayesian PAC-learning by \citep{Russo_Zou_2015,Xu_Raginsky_2017, pmlr-v83-bassily18a,Graepel05}. The generalization error is upper bounded by the mutual information between the training set and the learning algorithm output (i.e., the learned parameters if a parametric class model is allowed). More precisely, the learning process consists in mapping   training data to a particular model via a Markov kernel $p_{\hat{\theta}|\mathcal{S}_{n}}$. Then, the generalization error can be bounded as follows: 
\begin{equation}
\left|	\mathbb{E}_{p_{\mathcal{S}_{n}}p_{\hat{\theta} |\mathcal{S}_{n}}}\left[\mathcal{E}_{\textrm{gen-err}}({\mathcal{S}_n})\right] \right|\leq M_\theta\sqrt{\frac{1}{2n}\mathcal{I}\left(\hat{\theta}_{(\mathcal{S}_{n})} ; \mathcal{S}_{n} \right)},
\end{equation}
where $M_\theta$ is the maximum value of the loss function. However, in our present work, the framework, the underlying hypothesis and the tools are fundamentally different from Bayesian PAC-learning. Therefore, the  mutual information we obtained in this paper is between input features and latent representations,  which is very different than the above mutual information between the learned parameters and the training set. Our focus is to study the role of the mutual information between features and latent representations as a potential candidate to control regularization, as it was already shown to be useful in   \citep{2016arXiv161101353A, DBLP:journals/corr/AlemiFD016,8379424, Achille_Soatto_2017}, among others. Nonetheless,  the fact that two completely different approaches and models yield connections between mutual information and generalization confirms the importance of studying information-theoretic quantities in the context of statistical learning.



\section{Information-Theoretic Bounds on the Generalization Error}
\label{sec:mainbound}

In this section, we present our main result in Theorem~\ref{thm:regularizar}, which is a bound on the generalization error. In particular, we show that the mutual information between the input raw data and its representation controls the generalization with a scaling $\mathcal{O}\left(\frac{\log(n)}{\sqrt{n}}\right)$, which leads to a so-called informational generalization error bound. To this end, we will need some assumptions. 
{\begin{assumptions}\label{asumption1} Consider the following assumptions: 
	\begin{enumerate}
		\item Input space $\mathcal{X}\subset\mathbb{R}^{d_x}$ is closed and bounded, which implies that it has finite volume $\text{Vol}(\mathcal{X})<\infty$. Target space $\mathcal{Y}$ is finite $|\mathcal{Y}|<\infty$. In addition $\mathcal{X}\equiv \text{supp}(p_X)$ and $\displaystyle P_{Y}(y_{\min}) \coloneqq \min_{y\in\mathcal{Y}} P_Y(y)>0$. These are extremely mild conditions as we always can discard the sets of zero probability in $\mathcal{X}$ and $\mathcal{Y}$.
		\item Every encoder in the parametric class $q^\theta_{U|X}(u|\cdot)$ is continuous in $x$ for all $u\in\mathcal{U}\subset\mathbb{R}^{d_u}$, and its marginal pdfs has finite second order moment:
		\begin{align}
		\sup_{x\in\mathcal{X}} \, \sup_{\theta\in\Theta}\, \max_{j\in[1:d_u]}\mathbb{E}_{q^\theta_{U_j|X}}\left[U_j^2|X=x\right]\leq S<\infty,
		\end{align}
		where $u_j$ denotes the $j$-th entry in $u\in\mathcal{U}\subset\mathbb{R}^{d_u}$ with $j=1,\dots,d_u$.
		In addition, every decoder in the parametric model allocates non-zero probability mass:
		\begin{equation}
		Q_{\widehat{Y}|U}^\theta(y|u)>c, \qquad\forall\;\theta\in\Theta,\;u\in\mathcal{U},\;y\in\mathcal{Y},
		\end{equation}
		where $c>0$.
		\item The class of loss functions denoted by  $\Omega\coloneqq  \left\{\ell_\theta(x,y):\;\theta\in\Theta\right\}$ is totally bounded.
	\end{enumerate}
\end{assumptions}}

Theorem \ref{thm:regularizar} involves some steps which would worth explaining to facilitate the understanding of its statement. In particular, it uses two discretization procedures: one for the parametric space of the loss functions $\Omega$, which is needed to derive an uniform deviation of the generalization error (i.e., using the assumption that $\Omega$ is totally bounded), and another one for the input (feature) space $\mathcal{X}$, which is needed to introduce some information-theoretic measures such as the mutual information. In more precise terms: 

\begin{itemize}
	\item  The discretization of a parametric class of loss functions is a common procedure when we need a bound for the probability of uniform deviations with respect to an uncountable set. The typical approach is to bound the required probability using a worst-case criterion over a finite number of events with the help of the union bound. The use of {VC dimension} \citep{Devroye97a} of the underlying class or covering numbers  is the most common approach	in classical learning theory \citep{Devroye97a}. The later will be the approach we will follow here as well. In particular, we make use of the totally bounded hypothesis of $\Omega$, which allow us to guarantee that for all $\varepsilon>0$ there exists a finite parameter set $\mathcal{F}_{\Omega}^{(\varepsilon)}\coloneqq  \left\{\theta_i\right\}_{i=1}^{|\mathcal{F}_{\Omega}^{(\varepsilon)}|}\subset\Theta$ with $|\mathcal{F}_{\Omega}^{(\varepsilon)}|<\infty$ such that: for all $\theta\in\Theta$ (or $\ell_\theta\in\Omega$) there exists $i^\ast\in\{1,\dots,|\mathcal{F}_{\Omega}^{(\varepsilon)}|\}$ satisfying 
	\begin{equation}\label{eq:totallybounded}
	\sup_{x\in\mathcal{X}} \,\max_{y\in\mathcal{Y}} \, \left|{\ell}_{\theta}(x,y)-\ell_{\theta_{i^\ast}}(x,y)\right|<\varepsilon.
	\end{equation}
	\item The discretization of the input space $\mathcal{X}$ is used to  introduce in our problem an information-theoretic quantity such as the mutual information which is well  defined for finite alphabets spaces.  To this end, we introduce an artificial quantization of the input space following the approach in \citep{Xu2012}. Let us define, for each $y\in\mathcal{Y}$, a finite $\mathcal{B}_X$-measurable partition of the feature space $\mathcal{X}$ into $K$ connected cells $\{\mathcal{K}_k^{(y)}\}_{k=1}^K$ satisfying $\bigcup_{k=1}^K\mathcal{K}_k^{(y)}\equiv \mathcal{X},\;\mathcal{K}_{i}^{(y)}\cap\mathcal{K}_{j}^{(y)}=\emptyset,$ $\, \forall\, 1\leq i<j\leq K,\;\int_{\mathcal{K}_k^{(y)}}dx>0, \, \forall 1\leq k\leq K$. In addition,  let  $\{x^{(k,y)}\}_{k=1}^K$ be the respective cell centroids for each $y\in\mathcal{Y}$, so the partition family $\mathcal{K}$ is given by
	\begin{equation}\label{eq:partitionfamily}
	    \mathcal{K}=\left\{K,\big(\{\mathcal{K}_k^{(y)}\}_{k=1}^K,\{x^{(k,y)}\}_{k=1}^K\big)_{y\in\mathcal{Y}}\right\}.
	\end{equation}
	The partition $\mathcal{K}$ defines the cell radius (measured with respect to the encoder): 
	\begin{equation}\label{eq:bigdelta}
	    \Delta^\theta(K)=\sup_{\substack{1\leq k\leq K\\
	    	    (x,u,y) \in\mathcal{K}_k^{(y)}\times \mathcal{U}\times \mathcal{Y} }}\left|q^\theta_{U|X}(u|x)-q^\theta_{U|X}(u|x^{(k,y)})\right|.
	\end{equation}
Different possible partitions defines how small can these magnitudes be made. When $K$ increases, every cell $\mathcal{K}_k^{(y)}$ should naturally shrinks and the radius $\Delta^\theta(K)$ decreases (with the appropriate regularity conditions for the parametric class of encoders). The above discretization induces the following probability distribution for the input and output samples:
	\begin{equation}\label{eq:PxyD}
	P^{D}_{XY}(x,y)\coloneqq \sum_{k=1}^K\mathds{1}\Big\{x=x^{(k,y)}\Big\}P_{X|Y}(\mathcal{K}_k^{(y)}|y) P_Y(y).
	\end{equation}
	This probability distribution defines a new probability measure from  $P^{D}_{XY} q^\theta_{U|X}$ and a modified loss function, which will be used in the proof of Theorem \ref{thm:regularizar}:
	\begin{equation}\label{eq:ellD}
	{\ell}^D_\theta(x,y) \coloneqq  \mathbb{E}_{q_{U|X}^\theta}\left[\left.-\log Q_{{Y}|U}^{D,\theta}(y|U)\right|X=x\right],
	\end{equation}
where
	\begin{equation}\label{eq:Q_Y|U^D}
	    Q_{{Y}|U}^{D,\theta}(y|u)=\frac{\sum_{k=1}^KP_{XY}^D(x^{(k,y)},y)q^\theta_{U|X}(u|x^{(k,y)})}{\sum_{y^\prime\in\mathcal{Y}}\sum_{k=1}^KP_{XY}^D(x^{(k,y^\prime)},y^\prime)q^\theta_{U|X}(u|x^{(k,y^\prime)})}.
	\end{equation}
From the above definitions, it is understood implicitly  that the partition of $\mathcal{X}$ is independent on the parametric class of encoders. However, the value of $\Delta^\theta(K)$ is clearly a function of $\theta$. For different values of $\theta$ this number can be different. However, if the class of the encoders is sufficiently well-behaved, this variation will not be very wild. 
It is important to mention that the last discretization procedure is only used in the proof of the next theorem, but it does not restrict the validity of the result for continuous inputs probability distributions.
The reason for this is the monotonicity of the mutual information with respect to finite measurable partitions of the input space~(the reader is referred to \citep{pinsker1964information} for further details).

	
\end{itemize}

{ Now we are able to present our main result represented in a simple way in \eqref{eq:teorema_basicaversion}.} 

\begin{thm}[Main result]\label{thm:regularizar}
	For every $\delta\in(0,1)$, there exists a parameter set $\mathcal{F}_{\Omega}^{(\varepsilon)}\coloneqq\ \left\{\theta_i\right\}_{i=1}^{|\mathcal{F}_{\Omega}^{(\varepsilon)}|}\subset\Theta$ with $|\mathcal{F}_{\Omega}^{(\varepsilon)}|<\infty$ such that:
	\begin{equation}
	\mathbb{P}\left(|\mathcal{E}_{\text{gen-err}}(\mathcal{S}_n)|\leq\inf_{\varepsilon>0}\left\{\sup_{\theta\in \mathcal{F}_{\Omega}^{(\varepsilon)}}\mathcal{B}\left(\theta,\frac{\delta}{|\mathcal{F}_{\Omega}^{(\varepsilon)}|}\right)+2\varepsilon\right\}\right)\geq 1-\delta,
	\end{equation}
	where 
	\begin{align}
	\mathcal{B}(\theta,\delta)=&\inf_{\mathcal{K}}\left\{2\epsilon(\mathcal{K})+  r(\mathcal{K})	\mathrm{A}_\delta \sqrt{\mathcal{I}\big( U^\theta_{(X)};X  \big)}\cdot \frac{\log(n)}{\sqrt{n}}\right.\nonumber\\
	&\left.+\frac{\displaystyle\inf_{\beta>0}\frac{2e^{-1}g^\theta(\beta)(1+\beta)}{\beta}\left[\sqrt{2}r(\mathcal{K})\mathrm{B}_\delta\sqrt{\mathcal{I}\big( U^\theta_{(X)};X  \big)}\right]^{\frac{1}{1+\beta}}}{\sqrt{n}}\right\}\nonumber\\
	&+\frac{\mathrm{C}_\delta+\mathrm{D}_\delta\cdot \sqrt{\mathbb{E}_{p_{XY}}\left[T^\theta(X,Y)^2\right]}}{\sqrt{n}}+\mathcal{O}\left(\frac{\log(n)}{n}\right)\label{eq:bounderrorgap}
	\end{align}	
	with 
	\begin{align}
	\mathrm{A}_\delta& \coloneqq\sqrt{2}\mathrm{B}_\delta, \ \ \mathrm{B}_\delta \coloneqq 1+\sqrt{\log\left(\frac{|\mathcal{Y}|+4}{\delta}\right)} , \\
	\mathrm{C}_\delta& \coloneqq \log \left(\frac{(4\pi eS)^{d_u/2}}{P_{Y}(y_{\min})}\right)\sqrt{|\mathcal{Y}|}\mathrm{B}_\delta, \ \ 
	\mathrm{D}_\delta \coloneqq \sqrt{\frac{|\mathcal{Y}|+4}{\delta}},
	\end{align}
	and 
		\begin{align}
	g^\theta(\beta)  &\coloneqq\sup_{x,z\in\mathcal{X}}\sqrt{\int_\mathcal{U}q_{U|X}^\theta(u|x)\left(q_{U|X}^\theta(u|z)\right)^{\frac{-2\beta}{1+\beta}}du},\\
		T^\theta(x,y)  &\coloneqq\ell_\theta(x,y)-\tilde{\ell}_\theta(x,y),\label{eq:defgtrue}\\
\epsilon(\mathcal{K})&\coloneqq \sup_{\substack{x,k,y,\theta:\\1\leq k\leq K\\y\in\mathcal{Y}\\x\in\mathcal{K}_k^{(y)}\\\theta\in\Theta}}\left|\tilde{\ell}_\theta(x,y)-{\ell}^D_\theta(x^{(k,y)},y)\right|,\\
 r(\mathcal{K})&\coloneqq \frac{1}{\displaystyle\min_{\substack{k,y:\\1\leq k\leq K\\y\in\mathcal{Y}}}P_X\left(\mathcal{K}_k^{(y)}\right)}.\label{eq:epsilon_r_definitions}
	\end{align}
\end{thm}

The proof is relegated to Appendix~\ref{sec:proof_main_theorem}. This bound has some important terms which are worth analyzing:
\begin{itemize}
	\item $\mathcal{I}(U_{(X)}^\theta ; X )$: Mutual information between raw data $X$ and its randomized representation $U_{(X)}$ appears to be related to the generalization capabilities and thus to overfitting. It was interpreted as a ``measure of information complexity'' \citep{2016arXiv161101353A,DBLP:journals/corr/AlemiFD016,8379424}. Theorem  \ref{thm:regularizar} is a first step in order to explain how and why this effect happens. This term presents a scaling rate of $n^{-1/2}\log(n)$ and it is the main term in the generalization bound. It is well-known \citep{Devroye97a} that there exists bounds showing that the generalization error vanishes faster with $n$, but this intermediate regime appears to be visible in practical scenarios and these  results can  capture the dynamic of the generalization error with respect to some of hyperparameters regardless of this not optimal scaling. In Section \ref{sec:experimentos}, we present an empirical analysis that supports this claim.
	
	\item $\mathbb{E}_{p_{XY}}\left[T^\theta(X,Y)^2\right]$: This term can be interpreted as a measure of the decoder efficiency. It is basically, the mean-square error  between the loss function \eqref{eq:loss} and the modified loss $\tilde{\ell}_\theta(x,y)$ \eqref{eq:optimal_decoder} induced solely by the encoder. This magnitude can be also  understood as a measure of similarity between the $Q^\theta_{\widehat{Y}|U}$ and  the decoder $Q^{\theta}_{Y|U}$ induced by the encoder: 	
	\begin{equation}
	T^\theta(x,y)= \mathbb{E}_{q^\theta_{U|X}}\left[\left.\log\ \frac{Q^{\theta}_{Y|U}(y|U)}{Q^\theta_{\widehat{Y}|U}(y|U)} \right|X=x\right].
	\end{equation}
{ When $Q^\theta_{\widehat{Y}|U}=Q^{\theta}_{Y|U}$ this term is zero, suggesting that this selection could have a beneficial effect on the generalization error. This result is consistent with the bottleneck behaviour \eqref{eq:efectoib}: when this decoder is selected, the generalization error is controlled mainly by $\mathcal{I}(U^\theta_{(X)};X)$. 
When the selection of the decoder $Q^\theta_{\widehat{Y}|U}$ is close to $Q^{\theta}_{Y|U}$ (according to $	T^\theta(x,y)$) and the number of samples $n$ is such that $1/\sqrt{n}$ is considerably lower than $\frac{\log(n)}{\sqrt{n}}$, the mutual information $\mathcal{I}\big( U^\theta_{(X)};X  \big)$ has the main influence in the bound.}
	
	\item $\epsilon(\mathcal{K})$ and $r(\mathcal{K})$: The trade-off between these two magnitudes is obviously related to the original trade-off between $K$ and $\Delta^\theta(K)$ in \eqref{eq:bigdelta}. Clearly, $\epsilon(\mathcal{K})$ increases with $\Delta^\theta(K)$ (which is expected to decrease with $K$), i.e., if the encoders are close in the sense of \eqref{eq:bigdelta}, the decoders $Q^\theta_{Y|U}$ and $Q^{\theta,D}_{Y|U}$ will necessarily be. Similarly, $r(K)$ increases with $K$ (smaller cells contain less probability). As the exact trade-off is highly dependent, not only on the encoder parametric class, but also with respect to the exact input distribution, is not easy to obtain its accurate description. However, under some mild extra assumptions some analysis can be carried out, as showed in Section \ref{sec:particion}.
	\item $\frac{g^\theta(\beta)(1+\beta)}{\beta}$: This term depends entirely on the encoder. On the one hand, $\beta\mapsto \frac{1+\beta}{\beta}$ is a decreasing function, i.e.  $\frac{1+\beta}{\beta}\rightarrow\infty$ when $\beta\rightarrow0^+$ and $\frac{1+\beta}{\beta}\rightarrow1$ when $\beta\rightarrow\infty$. On the other hand, $g^\theta(\beta)$ is an increasing function such that $g^\theta(\beta)\rightarrow1$ when $\beta\rightarrow0^+$ and $g^\theta(1)\geq\sqrt{\text{Vol}(\mathcal{U})}$. So, when $\mathcal{U}$ is not bounded,  $\beta$ should be limited to $(0,1)$. In any case, this does not seem to be critical if the encoders class is chosen carefully, e.g., normal or log-normal distributions for which  this term can be shown to be finite provided that  $\beta\in(0,1)$.
\end{itemize}

\section{Assessing the Rationale of the Assumptions in Theorem \ref{thm:regularizar}}

\subsection{About the Total-Boundedness Hypothesis of $\Omega$}
\label{sec:casogaussiano}

A critical requirement for Theorem \ref{thm:regularizar} is the total-boundedness hypothesis on $\Omega$. This hypothesis allows us to analyze the generalization error using a worst-case (over the $\varepsilon$-net $F_\Omega^{(\varepsilon)}$) criterion  for the generalization error \eqref{eq:errorgap}. It is important to verify if this hypothesis can hold in practical scenarios. The goal of this section is to show that this is indeed the case. In order to show our claim, we will consider a common and popular choice of parametric encoders/decoders class. For example, for the class of encoders we will consider the case of Gaussian encoders \citep{2016arXiv161101353A,kingma2013auto}. That is,
\begin{equation}
q_{U|X}^{\theta_E}(u|x)=\prod_{i=1}^{d_u} \mathcal{N}\left(\mu_i(x,\beta_i),\sigma_i^2(x,\alpha_i)\right),
\end{equation}
where $\mathcal{N}(\mu,\sigma^2)$ denotes a Gaussian pdf with mean $\mu$ and variance $\sigma^2$. Functions $\mu_i(x,\beta_i)$ and $\sigma_i^2(x,\alpha_i)$ with $i=1,\dots, d_u$ are typically deep feed-forward neural nets, where the parameters $\theta_E\equiv \left\{\alpha_i,\beta_i\right\}_{i=1}^{d_u}$ are learned by minimizing the empirical risk \eqref{eq:empirical_risk}. We will assume that $\alpha_i,\beta_i\in\mathbb{R}^{l}$ for $i=1,\dots,d_u$. With these definitions it is clear that the total set of parameters for the encoder, that is $\theta_E$, lives in $\Theta_E\subseteq\mathbb{R}^{2ld_u}$. Note that with this choice for the encoder, $\mathcal{U}=\mathbb{R}^{d_u}$.

For the decoder parametric class we will consider the well-known soft-max architecture:
\begin{equation}
Q_{\hat{Y}|U}^{\theta_D}(k|u)=\frac{\exp\left\{\langle w_k,u\rangle+b_k\right\}}{\sum_{i=1}^{|\mathcal{Y}|}\exp\left\{\langle w_i,u\rangle+b_i\right\}},\ \ k=1,\dots,|\mathcal{Y}|,
\label{eq:soft_max_dec}
\end{equation} 
where $\theta_D\equiv \left\{w_i,b_i\right\}_{i=1}^{|\mathcal{Y}|}$ and $w_i\in\mathbb{R}^{d_u}$, $b_i\in\mathbb{R}$ for $i=1,\dots, |\mathcal{Y}|$ are the parameters of the soft-max chosen also by the minimization of the empirical risk. We clearly see that the set of decoder parameters ($\theta_D$) is in $\Theta_D\subseteq\mathbb{R}^{|\mathcal{Y}|(d_u+1)}$.

With these definitions, we can rewrite the set $\Omega$ as:
\begin{equation}
\Omega=\left\{\mathbb{E}_{q_{U|X}^{\theta_E}}\left[\left.-\log Q_{\hat{Y}|U}^{\theta_D}(k|U)\right|X=x\right]: (\theta_E,\theta_D)\in \Theta_E\times\Theta_D\right\}.
\label{eq:Omega_set_part}
\end{equation}

{Now we are ready to present the main result of this section. It requires very mild assumptions (see Assumptions \ref{asumption2}), usually satisfied in practice, on the parameter set $\Theta_E\times\Theta_D$ and on the boundedness and smoothness (Lipschitz continuity) of functions $\mu_i(x,\beta_i)$ and $\sigma_i^2(x,\alpha_i)$ with $i=1,\dots, d_u$. These assumptions guarantee the application of well-known results in functional analysis to prove the total-boundedness of $\Omega$. The above mentioned assumptions and the proof of the next theorem can be consulted in Appendix \ref{sec:proof_totbound}.}

\begin{thm}[Total-boundedness of $\Omega$]\label{thm:totbound}
	Under the set of Assumptions \ref{asumption2}, $\Omega$ in (\ref{eq:Omega_set_part}) is totally bounded.
\end{thm}
\begin{rem}
	Notice that the result of this theorem is valid for Gaussian encoders. However, it is not difficult to show that for other well-behaved encoders (e.g., log-normal ones \citep{2016arXiv161101353A}), the result also holds true. However, additional efforts are needed to show it. 
\end{rem}

\subsection{Some Basic Results on the Trade-off Between $\epsilon(\mathcal{K})$ and $r(\mathcal{K})$}\label{sec:particion}

In this section, we present two very simple lemmas that allow us to have a first glimpse on the trade-off between $\epsilon(\mathcal{K})$ and $r(\mathcal{K})$ and give some basic result on the scaling with the cell number $K$. 

\begin{lemma}\label{lem:epsilonDelta}
For every partition family $\mathcal{K}$, $\epsilon(\mathcal{K})\leq a_K$ with $a_K=\mathcal{O}(\sup_{\theta\in\Theta}\Delta^\theta(K))$.    
\end{lemma}
\begin{proof}:
From the definition of $\epsilon(\mathcal{K})$ in \eqref{eq:epsilon_r_definitions}, we can write:
\begin{equation}\label{eq:epsilonvound}
\epsilon(\mathcal{K})\leq\sup_{\theta\in\Theta}\left\{ \epsilon_0^\theta(K)+\max_{\substack{k,y:\\1\leq k\leq K\\y\in\mathcal{Y}}}\mathbb{E}_{q_{U|X}^\theta}\left[\left.\left|\log\frac{Q_{Y|U}^{D,\theta}(U|y)}{Q^\theta_{Y|U}(U|y)}\right|\right|X=x^{(k,y)}\right]\right\},
\end{equation}
where
\begin{align}
{\epsilon}_0^\theta(K)&=\sup_{(k,x,y)\in\mathcal{C}}\left|\tilde{\ell}_\theta(x,y)-\tilde{\ell}_\theta(x^{(k,y)},y)\right|
\end{align}
and $\mathcal{C}=\{(k,x,y):\;1\leq k\leq K,\;y\in\mathcal{Y},\;x\in\mathcal{K}_k^{(y)}\}$. It is easy to check, from the continuity of $\tilde{\ell}_\theta(x,y)$ with respect to $q_{U|X}(\cdot|x)$, that ${\epsilon}_0^\theta(K)=\mathcal{O}(\Delta^\theta(K))$. Something similar happens to the second term of \eqref{eq:epsilonvound}, where for every $u\in\mathcal{U}$ and $y\in\mathcal{Y}$:
\begin{align}
\log Q^\theta_{Y|U}(y|u)&=\log \frac{\sum_{k=1}^K\int_{\mathcal{K}_k^{(y)}}p_{XY}(x,y)q^\theta_{U|X}(u|x)dx}{\sum_{y^\prime\in\mathcal{Y}}\sum_{k=1}^K\int_{\mathcal{K}_k^{(y^\prime)}}p_{XY}(x,y^\prime)q^\theta_{U|X}(u|x)dx}\\
&\leq\log\frac{\sum_{k=1}^KP_{XY}^D(k,y)\left(q^\theta_{U|X}(u|x^{(k,y)})+\Delta^\theta(K)\right)}{\sum_{y^\prime\in\mathcal{Y}}\sum_{k=1}^KP_{XY}^D(k,y^\prime)\left(q^\theta_{U|X}(u|x^{(k,y^\prime)})-\Delta^\theta(K)\right)}.
\end{align}
Then, it is not hard to verify  that:
\begin{align}
\left|\log\frac{Q_{Y|U}^{D,\theta}(u|y)}{Q^\theta_{Y|U}(u|y)}\right|&=\mathcal{O}(\Delta^\theta(K)),
\end{align}
from which we can conclude that there exists $a_K=\mathcal{O}(\sup_{\theta\in\Theta}\Delta^\theta(K))$ such that $\epsilon(\mathcal{K})\leq a_K$.
	\begin{flushright}\vspace{-0.4cm}\fbox{\phantom{\rule{.1ex}{.1ex}}}\end{flushright}
\end{proof}

{When $K$ increases, $\Delta^\theta(K)$ decreases and Lemma \ref{lem:epsilonDelta} shows that $\epsilon(\mathcal{K})$ tends to decrease as well. The following lemma studies the trade-off between  $r(\mathcal{K})$ and $\epsilon(\mathcal{K})$ under rather  reasonable assumptions.
\begin{lemma}\label{lem:scaling}
Let  $q^\theta_{U|X}(u|\cdot)$ be a parametric class of Lipchitz continuous encoders and   for every partition family $\mathcal{K}$:
\begin{equation}
	\min_{\substack{k,y:\\1\leq k\leq K\\y\in\mathcal{Y}}} \mathbb{P}\left(X\in\mathcal{K}_k^{(y)}\right)\geq b_K,\quad\max_{\substack{k,y:\\1\leq k\leq K\\y\in\mathcal{Y}}}\text{Vol}(\mathcal{K}_k^{(y)})\leq c_K \label{eq:scale_cond}
	\end{equation}
with $b_K=\mathcal{O}(K^{-1})$ and $c_K=\mathcal{O}(K^{-1})$. Then, $r(\mathcal{K})=\mathcal{O}(K)$, $\Delta^\theta(K)=\mathcal{O}\left(K^{-1/d_x}\right)$ and $\epsilon(\mathcal{K})=\mathcal{O}\left(K^{-1/d_x}\right)$.
\end{lemma}
\begin{rem} The conditions required in Lemma \ref{lem:scaling}  are not highly  restricting. Notice that it is possible to show that
\begin{equation}
\min_{\substack{k,y:\\1\leq k\leq K\\y\in\mathcal{Y}}} \mathbb{P}\left(X\in\mathcal{K}_k^{(y)}\right)\leq\frac{1}{K},\quad\max_{\substack{k,y:\\1\leq k\leq K\\y\in\mathcal{Y}}}\text{Vol}(\mathcal{K}_k^{(y)})\geq\frac{\text{Vol}(\mathcal{X})}{K}
	\end{equation}
with equalities in equiprobable and equivolume partitions, respectively. As a consequence,  it would be reasonable that given good partitions it can be found for \eqref{eq:bounderrorgap} that the above quantities  do not deviate significantly from the these behaviours.	
\end{rem}
\begin{proof}
The proof of $r(\mathcal{K})=\mathcal{O}(K)$ is an immediate consequence of \eqref{eq:scale_cond}. From Lipschitz continuity of $q^\theta_{U|X}(u|\cdot)$ in $\mathcal{K}_k^{(y)}$ and the relationship with the volume, it is easy to show that 
\begin{align}
\Delta^\theta(K)&=\mathcal{O}\left(\sup_{(k,x,y)\in\mathcal{C}}\|x-x^{(k,y)}\|_2\right)=\mathcal{O}\left(V_M(K)^{1/d_x}\right),   
\end{align} 
where
\begin{equation}
V_M(K)=\max_{\substack{k,y:\\1\leq k\leq K\\y\in\mathcal{Y}}}\text{Vol}(\mathcal{K}_k^{(y)})=\mathcal{O}(K^{-1}). 
\end{equation}
Then, $\Delta^\theta(K)=\mathcal{O}\left(K^{-1/d_x}\right)$ and using Lemma \ref{lem:epsilonDelta}, $\epsilon(\mathcal{K})=\mathcal{O}\left(K^{-1/d_x}\right)$.
	\begin{flushright}\vspace{-0.4cm}\fbox{\phantom{\rule{.1ex}{.1ex}}}\end{flushright}
\end{proof}
}

\section{Experimental Results}
\label{sec:experimentos}

The goal of our experiments in this section is to validate the results from Theorem \ref{thm:regularizar}, by showing that indeed the mutual information is representative of the generalization error, at least in a qualitative fashion. In other words, by controlling the mutual information between features and representations we aim at investigating if we can better control generalization in  architectures of limited capacity. The magnitudes are compared for several rules $(q_{U|X}^\theta,Q^\theta_{\hat{Y}|U})$ considering also the influence of the Lagrange multiplier used in each experiment for controlling the level of regularization during training. 
\begin{rem}
Theorem \ref{thm:regularizar} studies the link between generalization error and mutual information for a given $\theta$, which yields a worst-case bound over a finite parameter set $\mathcal{F}_\Omega^{(\varepsilon)}$. As this set cannot be known in practice, we are going to make the comparison for the $\theta$ found by the algorithm during training stage. Since this worst-case criterion is used for theoretical convenience to find inequalities and that the comparison to be made is merely qualitative, this decision does not seem far-fetched and maintains the spirit of the bound.  
\end{rem}

There exists the difficulty of implementing a mutual information estimator. In practice we have a product-form encoder:  $q^\theta_{{U}|{X}}({u}|{x})=\prod_{j=1}^{d_u}q^\theta_{U_j|\mathbf{X}}(u_j|\mathbf{x})$ but  the marginal distribution $q^\theta_{{U}}({u})=\mathbb{E}_{p_X}\left[q^\theta_{U|{X}}({u}|{X})\right]$ does not necessarily satisfy this property. To this end, we make use of a variational bound \citep{cover} for  mutual information as follows:
\begin{align}
\mathcal{I}\left(U^\theta_{(X)}; X\right) &= \mathbb{E}_{p_{{X}}} \left[ \textrm{KL}\left(q^\theta_{{U}|{X}} \big\| \tilde{q}^\theta_{{U}} \right)\right]-\textrm{KL}(q^\theta_{{U}}\|\tilde{q}^\theta_{{U}})\\
&\leq\sum_{j=1}^{d_u}\mathbb{E}_{p_{{X}}} \left[ \textrm{KL}\left(q^\theta_{U_j|{X}}(\cdot|{X}) \big\| \tilde{q}^\theta_{U_j}  \right)\right],
\label{eq:inf_radius_opt}
\end{align}
where $\tilde{q}^\theta_{{U}}({u})=\prod_{j=1}^{d_u}\tilde{q}^\theta_{U_j}(u_j)$ is an auxiliary prior pdf \citep{kingma2013auto,2016arXiv161101353A} which may or not depend on $\theta$. The best choice in the above inequality is $\tilde{q}^\theta_{U_j}(u_j)=\mathbb{E}_{p_{{X}}}\left[q^\theta_{U_j|{X}}(u_j|{X})\right],
$ $j=[1:d_u]$. 

\begin{table}[t]
	\centering
	\begin{tabular}{ |c||c|c|c|c| }
		\hline 
		&$q^\theta_{U_j|\mathbf{X}}$&$Q^\theta_{\hat{Y}|\mathbf{U}}$&$\tilde{q}^\theta_{U_j}$&Motivated by\\   
		\hline
		Ex. 1 & Normal & Softmax & $\mathcal{N}(0,1)$ & \citep{kingma2013auto}\\
		\hline
		Ex. 2 & Log-Normal & Softmax & Log-Normal &\citep{2016arXiv161101353A}\\
		\hline
		Ex. 3 & RBM & Softmax & $\frac{1}{n}\sum_{i=1}^nq_{U_j|\mathbf{X}}^\theta(u_j|\mathbf{x}_i)$ &\citep{Hinton_guia}\\
		\hline
	\end{tabular}
	\caption{Architectures to be implemented.}
	\label{tabla}
\end{table} 
We will make use of a parametric estimator of the KL divergence based on: 
\begin{equation}\label{eq:klaprox}
\mathbb{E}_{p_{{X}}} \left[ \textrm{KL}\left(q^\theta_{U_j|{X}}(\cdot|{X}) \big\| \tilde{q}^\theta_{U_j}  \right)\right]\approx\frac{1}{n}\sum_{i=1}^n\textrm{KL}\left(q^\theta_{U_j|{X}}(\cdot|x_i) \big\| \tilde{q}^\theta_{U_j} \right),
\end{equation}

 We refer to each of the examples by the specifics of its encoder: Example $1$ uses a  Normal encoder while example $2$ uses a Log-Normal one and example $3$ a RBM, as can be seen in Table \ref{tabla}. Note that the mutual information does not depend on the decoder, so for simplicity we use always a \emph{soft-max} output layer. The values reported in each simulation are the average of three independent simulations, choosing at random and in each case, different sets for training and testing.

As our main goal is not to present a new classification architecture, with competitive  state-of-the-art methods, we restrict ourself to small subsets of databases, as motivated by \citep{Neyshabur2017GeometryOO}. More specifically, we  sample two different random subsets of: MNIST (standard dataset of handwritten digits) and CIFAR-10 (natural images \citep{cifar}). The size of the training and testing set are $5K$ and $10K$ respectively for both datasets. It is important to emphasize the main difference between the datasets:  MNIST proposes a task that is simpler than CIFAR-10, especially in presence of a small number of samples and without convolutional networks. From this observation, we deal with two different regimes : one in which high accuracy is achieved and another in which the algorithm is unable to achieve good performances, as can be seen in Table \ref{acuracy}. That is, we expect that our main result be valid not only in situations where the final accuracy is good enough, but also when final classification performance is not so good.  In addition, in Appendix \ref{app:experimentos} we present some complementary experiments.

\begin{rem}
RBM encoder assumes a discrete latent variable $U$ while our main result is for continuous random variables. Theorem \ref{thm:regularizar} can be modified, by relying on  tools from \citep{Shamir:2010:LGI:1808343.1808503, ourisit18} for discrete alphabets. 
\end{rem}

\begin{table}[t]
	\centering
	\begin{tabular}{ |c||c|c|c| }
		\hline 
		&Normal&Log-Normal&RBM\\   
		\hline
		MNIST & $0.951$ & $0.933$ & $0.936$\\
		\hline
		CIFAR-10 & $0.429$ & $0.446$ & $0.398$\\
		\hline
	\end{tabular}
	\caption{Best accuracy achieved with each architecture.}
	\label{acuracy}
\end{table} 

\subsection{Normal Encoder: Variational Classifier}\label{sec:normal_encoder}

Gaussian Variational Autoencoders (VAEs) introduce a conditionally independent normal encoder $U_j|_{{X}={x}}\sim\mathcal{N}(\mu_j({x}),\sigma^2_j({x}))$, $j=[1:d_u]$, where $\mu_j({x})$ and $\log\sigma_j^2({x})$ are constructed via deep neural networks, a standard normal prior $\tilde{U}_j\sim\mathcal{N}(0,1)$ and the decoder input is generated by sampling based on the well-known reparameterization trick \citep{kingma2013auto}. Each KL divergence involved in expression \eqref{eq:klaprox} can be computed as
\begin{equation}
\textrm{KL}\left(q^\theta_{U_j|{X}}(\cdot|x_i) \big\| \tilde{q}^\theta_{U_j} \right)=\frac{1}{2}\left(-\log\sigma^2_j({x}_i)+\sigma_j^2({x}_i)+\mu_j^2({x}_i)-1\right).
\end{equation} 
We consider a deep neural network composed of a feed-forward layer of $512$ hidden units with ReLU activation followed by another linear one for each parameter ($\mu$ and $\log\sigma^2$) with $256$ hidden units. That is each parameter, $\mu$ and $\log\sigma^2$, are a two-layer network where the first one is common to both\footnote{The implementation of $\log\sigma^2$ instead of $\sigma^2$ prevents the numerical optimization from finding degenerate a normal of zero variance.}. We chose a learning rate of $0.001$, a batch-size of $100$ and we trained during $200$ epochs. The cost function considered during the training phase was of the form
\begin{equation}\label{eq:normal-lognormal cost function}
\mathcal{L}_{\text{emp}}(q^\theta_{U|X},Q^\theta_{\hat{Y}|U},\mathcal{S}_{n})+\lambda \sum_{j=1}^{d_u}\frac{1}{n}\sum_{i=1}^n\textrm{KL}\left(q^\theta_{U_j|{X}}(\cdot|x_i) \big\| \tilde{q}_{U_j} \right),
\end{equation} 
where $\lambda\geq 0$ is the regularizing Lagrange multiplier. This approach is known as  $\beta-$VAE \citep{higgins17}, which matches the classic VAE  when the Lagrange multiplier satisfies $\lambda=1$. A $\beta-$variational classifier was used by \citet{pmlr-v97-li19a} and \citet{maggipinto2020betavariational} among others. 

\begin{figure}
    \centering
    \includegraphics[width=\textwidth]{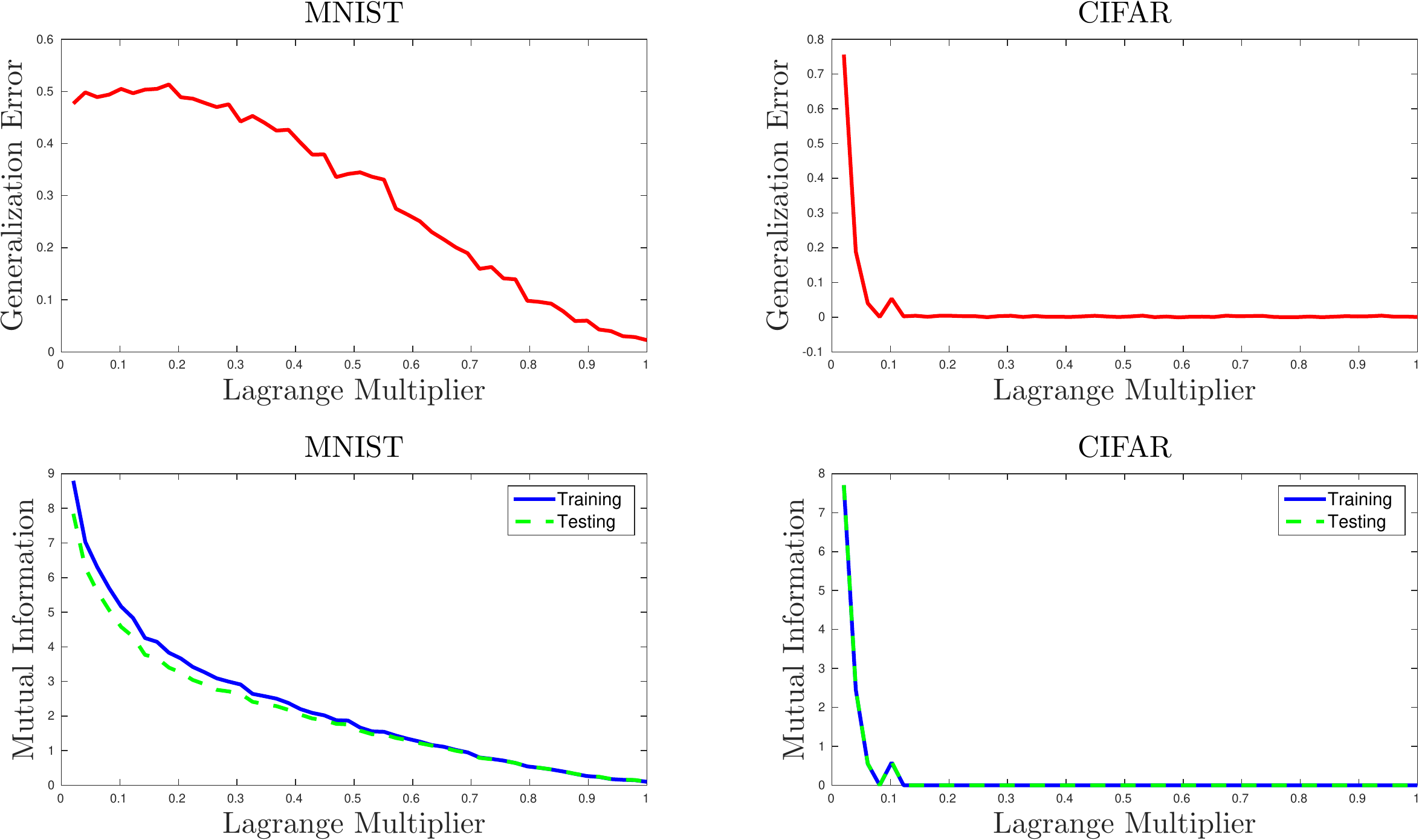}
    \caption{Generalization error and mutual information for a normal encoder architecture. Curves on the left correspond to experiments on the MNIST database and those on the right  correspond to CIFAR-10.}
    \label{fig:normal}
\end{figure}

Fig. \ref{fig:normal} shows generalization error and mutual information behaviour for experiments on MNIST and CIFAR-10 datasets. As the mutual information estimator is used as a regularization term, it is reasonable to expect a decreasing behavior of this with respect to the Lagrange multiplier. In addition, KL divergence is a classic regularization term for a normal encoder, with which it is also expected a decreasing behavior in terms of the generalization error as well. Both decreasing behaviors are not only experimentally corroborated in Fig. \ref{fig:normal}, but a certain similarity is seen in the way that both are decreasing, especially in our experiments with CIFAR-10. For the case of MNIST dataset the behavior becomes more similar as the Lagrange multiplier grows. It should also be added that the mutual information estimated with the training set and with the testing one are very much the same.  Therefore, it can give a qualitative notion of generalization error with respect to the Lagrange multiplier without the need of relying on a validation set.

\subsection{Log-Normal Encoder: Information Dropout}

{Information dropout proposes conditionally-independent log-normal encoders $U_j=f_j({X})e^{\alpha_j({X})Z}$, $j=[1:d_u]$ where $Z\sim\mathcal{N}(0,1)$, and $f_j({x})$, and $\alpha_j^2({x})$ are constructed via deep feed-forward neural networks and the decoder input is generated by sampling using the reparametrization trick. We follow the approach \citep{2016arXiv161101353A}, where it is recommended to use a log-normal prior $\tilde{U}_j\sim\log\mathcal{N}(\mu_j,\sigma_j^2)$ where $(\mu_j,\sigma_j)$ are parameters to be learned.}

Since the KL divergence is invariant under reparametrizations, the divergence between two log-normal distributions is equal to the divergence between the corresponding normal distributions \citep{cover}. Therefore, using the formula for the KL
divergence of normal random variables, we obtain 
\begin{align}
\textrm{KL}\left(q^\theta_{U_j|{X}}(\cdot|x_i) \big\| \tilde{q}^\theta_{U_j}  \right)&=\textrm{KL}\left(\log\mathcal{N}(\log f_j({x}_i),\alpha_j^2({x}_i))\|\log\mathcal{N}(\mu_j,\sigma_j^2)\right)\\
&=\textrm{KL}\left(\mathcal{N}(\log f_j({x}_i),\alpha_j^2({x}_i))\|\mathcal{N}(\mu_j,\sigma_j^2)\right)\\
&=\frac{\alpha_j^2({x}_i)+(\log(f_j({x}_i))-\mu_j)^2}{2\sigma_j^2}-\log\frac{\alpha_j({x}_i)}{\sigma_j}-\frac{1}{2}.
\end{align} 

For ${f}({x})=[f_1({x}),\cdots, f_{d_u}({x})]$ we used a feed-forward structure with two layers of $256$ hidden units with a softplus activation. We let $\alpha({x})=[\alpha_1({x}),\cdots,\alpha_{d_u}({x})]$ to be a feed-forward layer of $256$ hidden units with a sigmoid activation multiplied by $0.7$, so that the
maximum variance of the log-normal error distribution is approximately $1$ preventing null variances \citep{2016arXiv161101353A}. We chose a learning rate of $0.001$, a batch-size of $100$ and we trained during $200$ epochs. The cost function trained was the same than in \eqref{eq:normal-lognormal cost function}. 

\begin{figure}
    \centering
    \includegraphics[width=\textwidth]{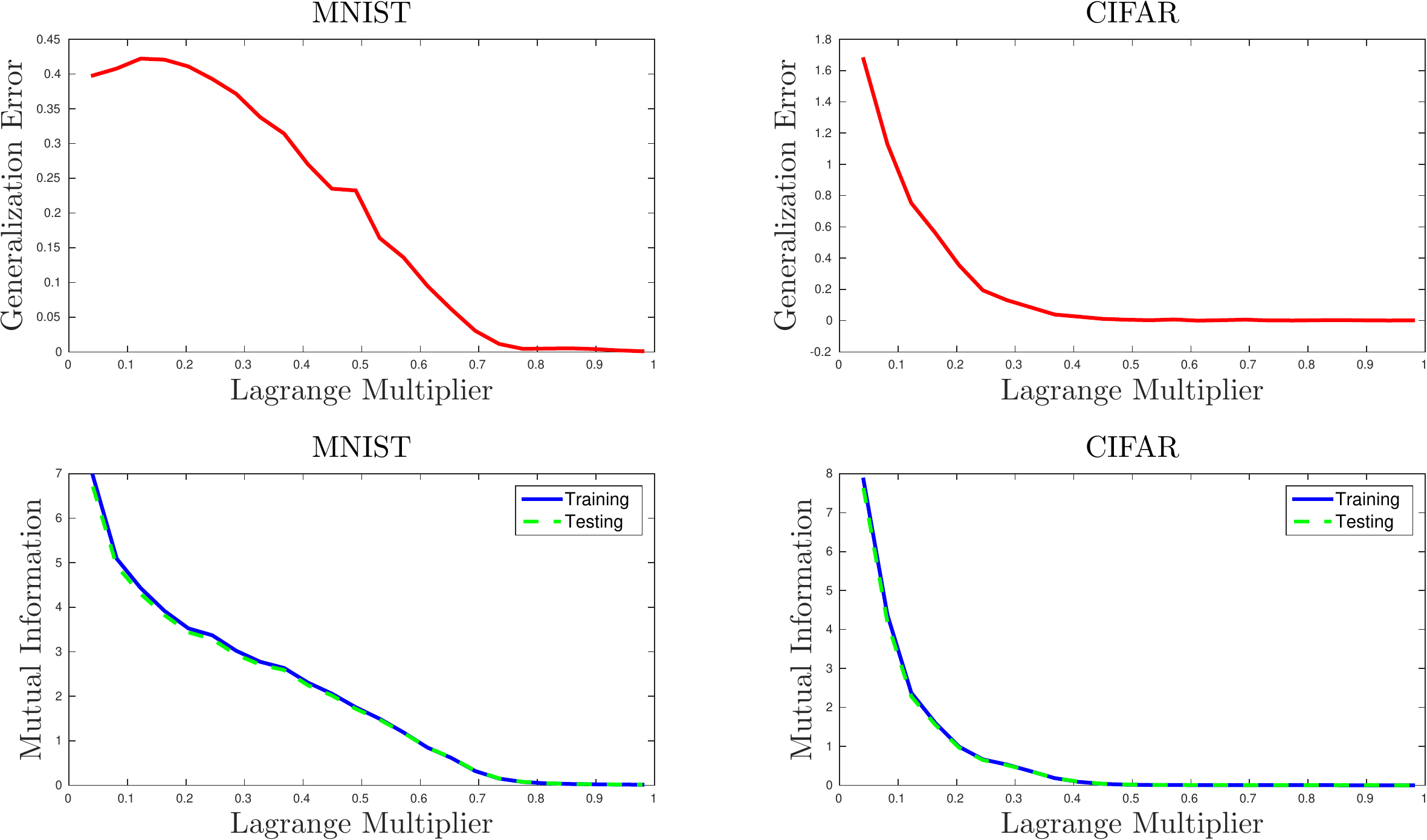}
    \caption{Generalization error and mutual information for a lognormal encoder architecture. Curves on the left correspond to experiments on the MNIST database and those on the right correspond to CIFAR-10.}
    \label{fig:lognormal}
\end{figure}

Fig. \ref{fig:lognormal} shows generalization error and mutual information behaviour for experiments on MNIST and CIFAR-10 datasets. As in the normal-encoder case, it is expected that both the mutual information and the generalization error have a decreasing behavior: in the first case because it is controlled by the Lagrange multiplier and in the second because this type of regularization is known to perform well for this architecture \citep{2016arXiv161101353A}. Both decreasing behaviors are not only experimentally corroborated in Fig. \ref{fig:lognormal}, but both curves have a similar shape in terms of decay. Again this similarity is more pronounced in experiments on CIFAR-10 and on MNIST for moderately high Lagrange multipliers. It is also seen that the mutual-information estimation with the training set is almost as good as with the testing one.

\subsection{RBM Encoder: Classification Using Restricted Boltzmann Machines}\label{sec:rbm_simulation}

In this section, we consider the standard models for RBMs studied in \citep{Hinton_guia,srivastava2014dropout}. For every $j\in[1:d_u]$, $U_j$ given ${X}={x}$ is distributed as a \emph{Bernoulli} random variable with parameter $\sigma(b_j+{w}_j^T{x})$ (sigmoid activation). {By selecting the prior distribution $\tilde{q}^\theta_{{U}}({u})=\prod_{j=1}^{d_u}\tilde{q}^\theta_{U_j}(u_j)$ with $
\tilde{q}_{U_j}(u_j)=\frac{1}{n}\sum_{i=1}^nq_{U_j|{X}}(u_j|{x}_i)
$ we obtain}
\begin{align}
\textrm{KL}&\left(q^\theta_{U_j|{X}}(\cdot|x_i) \big\| \tilde{q}^\theta_{U_j} \right)=\sigma(b_j+\langle{w}_j,{x}_i\rangle)\log\left(\frac{\sigma(b_j+\langle{w}_j,{x}_i\rangle)}{\frac{1}{n}\sum_{k=1}^n\sigma(b_j+\langle{w}_j,{x}_k\rangle)}\right)+\nonumber\\
&\qquad\qquad+\left(1-\sigma(b_j+\langle{w}_j,{x}_i\rangle)\right)\log\left(\frac{1-\sigma(b_j+\langle{w}_j,{x}_i\rangle)}{\frac{1}{n}\sum_{k=1}^n1-\sigma(b_j+\langle{w}_j,{x}_k\rangle)}\right).\label{eq:rbm}
\end{align}
Eq. \eqref{eq:rbm} is difficult to be used as a regularizer even with the contrastive divergence learning procedure by \citep{Hinton2002Training}. Instead, we rely on the usual RBM regularization: weight-decay. This is a traditional way to improve the generalization capacity. We explore the effect of the Lagrange multiplier $\lambda$, so called weight-cost, over both the generalization error and the mutual information. This meta-parameter controls the gradient weight decay, i.e., the cost function can be written as:
\begin{equation}
\textrm{CD}_{\text{RBM}}+\frac{\lambda}{2}\sum_{j=1}^{d_u}\|w_j\|_2^2,
\end{equation}
where $\textrm{CD}_{\text{RBM}}$ is the classical unsupervised RBM cost function trained via the contrastive divergence learning procedure by \citep{Hinton_guia}. In order to compute the generalization error we added to the output of the last RBM layer a soft-max regression decoder trained during $500$ epochs separately. Several authors have combined RBMs with soft-max regression  \citep{Hinton2006_DBN,DBLP:conf/uai/SrivastavaSH13,Chopra2018}, among others. 

Following suggestions from \citep{Hinton_guia}, we study the Lagrange multiplier when $\lambda\geq0.00001$ and plot the curves on a logarithmic scale. We fixed learning rate at  $0.1$, chose a batch-size of $100$, {and used $256$ hidden units.} We trained during $200$ epochs and started with a momentum of $0.5$ and change to $0.9$ after $5$ epochs. 

\begin{figure}
    \centering
    \includegraphics[width=\textwidth]{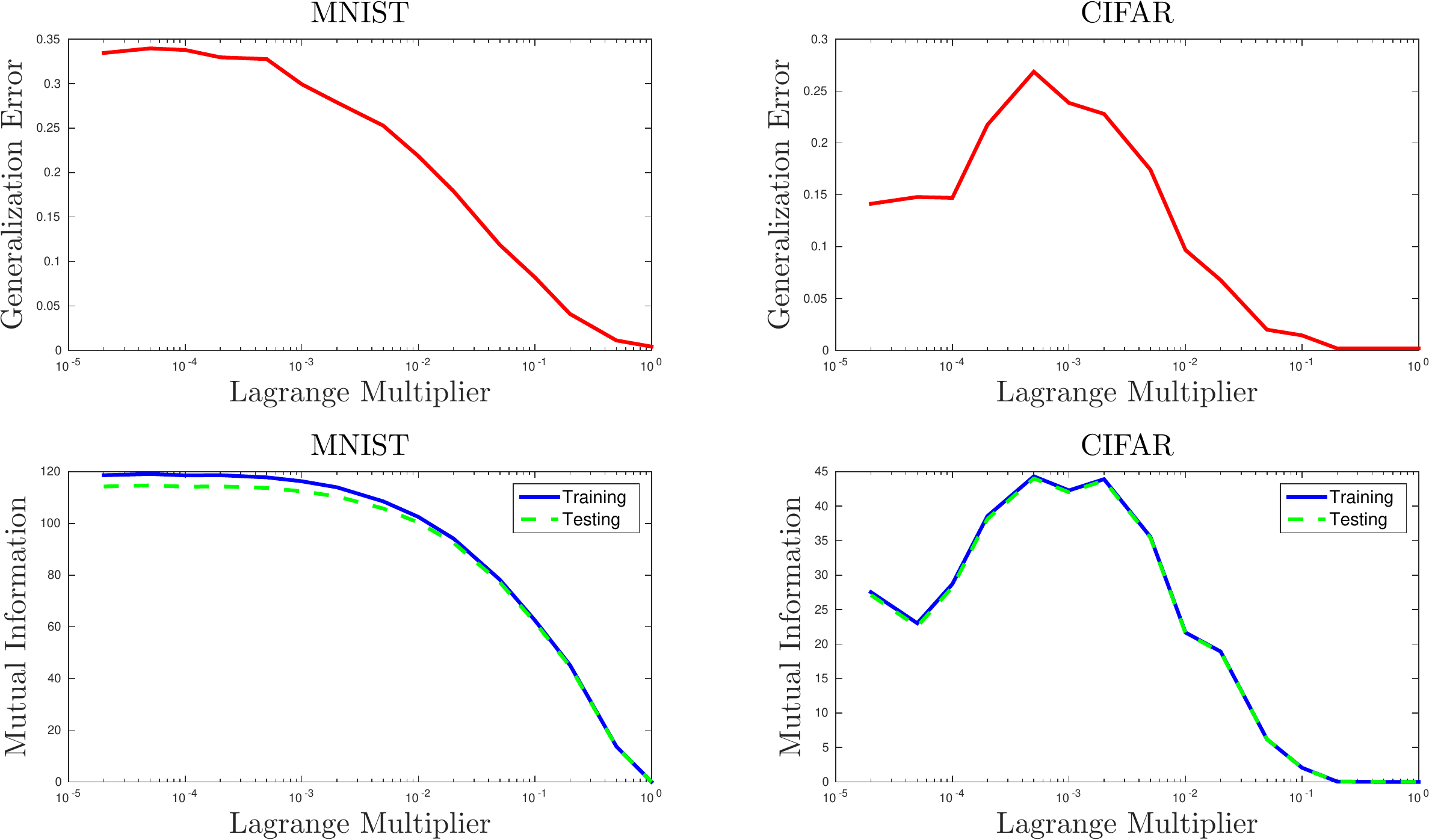}
    \caption{Generalization error and mutual information for a RBM encoder architecture. Curves on the left correspond to experiments on the MNIST database and those on the right correspond to CIFAR-10.}
    \label{fig:rbm}
\end{figure}

Fig. \ref{fig:rbm} shows generalization error and mutual information behaviour for experiments on MNIST and CIFAR-10 datasets. As weight decay is a standard regularization term, it is reasonable to expect a decreasing behavior of the generalization error. We observe a decreasing behavior of the mutual information which is close enough to that of the generalization error. Again, the mutual information estimator using the training set performs as good as the one that uses the testing set.

\section{Summary and Concluding Remarks}
\label{sec:conclusion}

In this work we presented a theoretical of the generalization error between the expected cross-entropy risk and its empirical approximation measured with respect to the training set. The  main result is stated in Theorem \ref{thm:regularizar} which shows that the generalization error can be upper bounded (with high probability) by two major terms which depend on the mutual information between the input and the corresponding latent representations (generated by a parametric class  of stochastic encoders in use), and a measure related to the decoder's efficiency, among others factors.  Our proof borrows tools from algorithm robustness \citep{Xu2012} introducing a discretization of the input (feature) space and the encoder parametric class.  Beside this, we provided formal support to show that our critical assumptions needed for the proof of Theorem \ref{thm:regularizar} can be easily fulfilled in practice by popular classes of encoders (cf. Section \ref{sec:casogaussiano}).

Experimental results on real-image datasets using well-known encoders models based on deep neural networks were  used to validate the existence of the statistical regime predicted by Theorem \ref{thm:regularizar} where the mutual information between input features and the corresponding latent representations allows to predict relatively well the behaviour of the generalization error with respect to some important structural parameters of the learning algorithm, e.g.,  the Lagrange multiplier weighting the regularization term in the cross-entropy loss. Of course, further numerical analysis is needed, in particular exploring  more  sophisticated deep neural networks models, but our results indicate that regularization by mutual information can have a direct influence in the generalization error. 

{A possible practical continuation of this work should be the development of regularization techniques based on the squared root of mutual information. The main obstacle for this goal is the availability of robust mutual information estimators in high dimensions. It would also be interesting, as a theoretical continuation, to analyze how tight the bound presented in this work is.} 





%
%

\appendix

\section{Proof of Theorem~\ref{thm:regularizar}} \label{sec:proof_main_theorem}

In this Appendix we will prove the main theorem of this work. We will use some well-known results, listed in Appendix \ref{app:auxiliary-results}. Before beginning some comments regarding the general approach to be followed in the proof are presented:
\begin{enumerate}
	\item Statistical dependence between the training set $\mathcal{S}_n$ and the parameters $\theta$ (or equivalent the encoder/decoder pair) is a major issue when trying to obtain bounds on the probability tails of the generalization error. The first step to avoid this complication is to use the totally bounded hypothesis about class loss $\Omega$  (item 4 of Assumptions \ref{asumption1}) to break-down the above mentioned dependency with a worst-case criterion. This is approach is common in the classical statistical learning theory.
	\item While our decision rule depends on jointly on the encoder and decoder, the mutual information term will only depends on the encoder. For this reason we will decouple the influence of the decoder in our bound in a different term.
	\item We will look for a relationship between error gap (to be defined in Section \ref{subsec:error_gap}) and mutual information, which is a \emph{information-theoretic} measure. Although mutual information is well-defined for continuous alphabets (under some regularity conditions), the case for discrete alphabets is very important and generally, easier to handle. This is the case in Shannon theory \citep{cover} and also in some results in learning theory.
For example in \citep{Shamir:2010:LGI:1808343.1808503}, authors bounds the deviation over the mutual information between labels $Y$ and hidden representations $U$, through the mutual information between hidden representations and inputs $X$ with discrete alphabets. Of course, through the use of variational representations, like the Donsker-Varadhan formula \citep{Donsker_Varadhan_1983}, mutual information for continuous alphabets can be easily introduced in learning problems \citep{Russo_Zou_2015,Xu_Raginsky_2017}. However, such results lead to mutual information terms between the inputs and the outputs of the learning algorithm. Although such results are very interesting, it is often very difficult to model the conditional distribution of the output of the algorithm given the input samples (which is needed for the full characterization of the mutual information term). Our approach here is different. Inspired by the information bottleneck criterion, we consider the effect on the generalization error, of the mutual information between the inputs $X$ and the the representations $U$, which are generated using the parametric class on encoders. An easier way to obtain such term is considering that the input space is discrete (as in \citep{Shamir:2010:LGI:1808343.1808503}). As this hypothesis is not easy to justify in a typical learning problem, we will achieve our desired result through a careful discretization of the input space and the use, in a final step, of the well-known Data Processing Inequality\footnote{As a matter of fact, the Data Processing Inequality is not needed. It suffices, to use the well-known monotonicty properties of f-divergences measures (of which the mutual information is a special case) \citep{pinsker1964information}.}. The approach used for the discretization of the input space has points in common with the \emph{robust-algorithms theory} \citep{Xu2012}.

\end{enumerate}

\subsection{Error-gap: A Worst-Case Bound for the Generalization Error}
\label{subsec:error_gap}

Our first step will be to bound the generalization error with a worst criterion using the hypothesis about $\Omega$ (item 4 of Assumptions \ref{asumption1}). 
\begin{lemma}\label{lem:coveringnet}
Under Assumptions \ref{asumption1}, let $\mathcal{F}_{\Omega}^{(\varepsilon)}=\left\{\theta_i\right\}_{i=1}^{|\mathcal{F}_{\Omega}^{(\varepsilon)}|}\subset\Theta$ with $|\mathcal{F}_{\Omega}^{(\varepsilon)}|<\infty$ the $\varepsilon$-net introduced in \eqref{eq:totallybounded} for $\varepsilon>0$. Let $\delta>0$ and $\alpha_\delta$ such that:
	\begin{equation}
	\max_{\theta\in \mathcal{F}_{\Omega}^{(\varepsilon)}}\mathbb{P}\left(\mathcal{E}_{\text{gap}}(f_\theta,\mathcal{S}_n)\leq\alpha_\delta-2\varepsilon\right)>1-\delta,
	\end{equation}
	where the \emph{error-gap} is defined as:
	\begin{equation}\label{eq:errorgap}
	\mathcal{E}_{\text{gap}}(f_\theta,\mathcal{S}_n)=\Big|\mathcal{L}\left(f_\theta\right)-\mathcal{L}_\text{emp}\left(f_\theta,\mathcal{S}_n\right)\Big|.
	\end{equation}
The generalization error is upper bounded, with probability at least $1-\delta$, with $\alpha_{\delta/|\mathcal{F}_\Omega^{(\varepsilon)}|}$, i.e.
	\begin{align}
	\mathbb{P}\left(\left|\;\mathcal{E}_{\text{gen-err}}(\mathcal{S}_n)\right|\leq\alpha_{\delta/|\mathcal{F}_\Omega^{(\varepsilon)}|}\right)\geq1-\delta.
	\end{align}
\end{lemma}

\begin{rem}
	Notice that we introduce what we have called the error gap, in order to provide a bound for the generalization error. 
	Notice that the error gap is defined for fixed parameters or independent from the training set. In this sense, the error gap is not the same that the generalization error but both concepts are related. This is similar to Vapnik's approach \citep[Chapter 12]{Devroye97a}, in which the generalization error can be bounded by a worst-case in the error gap:
	\begin{equation}\label{eq:vapnik}
	\left|\;\mathcal{E}_{\text{gen-err}}(\mathcal{S}_n)\right|\leq\sup_{\theta\in\Theta} \mathcal{E}_{\text{gap}}(f_\theta,\mathcal{S}_n).
	\end{equation} 
\end{rem}

\begin{proof}
	As $\Omega$ is totally bounded there exists a $\theta^\ast\in \mathcal{F}_{\Omega}^{(\varepsilon)}$ such that
	\begin{equation}
	\sup_{x\in\mathcal{X}}  \,\max_{y\in\mathcal{Y}}  \,  \left|\ell_{\widehat{\theta}_n}(x,y)-\ell_{\theta^\ast}(x,y)\right|<\varepsilon,
	\end{equation}
	where $\widehat{\theta}_n$ is the minimum over $\Omega$ of the empirical risk as defined in (\ref{eq:thetahatn}). Then, 
	\begin{align}
	\left|\mathcal{L}\left(
	f_{\widehat{\theta}_n} \right)-\mathcal{L}\left( f_{{\theta}^\ast}
	\right)\right|&\leq\varepsilon,\\
	\left|\mathcal{L}_\text{emp}\left( f_{\widehat{\theta}_n}, \mathcal{S}_n\right)-\mathcal{L}_\text{emp}\left(
		f_{{\theta}^\ast},\mathcal{S}_n\right)\right|&\leq\varepsilon.
	\end{align}
	The generalization error can be bounded using triangle inequality: 
	\begin{align}
	|\mathcal{E}_{\text{gen-err}}(\mathcal{S}_n)|&=\left|\mathcal{L}\left(  f_{\widehat{\theta}_n} \right)-\mathcal{L}_\text{emp}\left(
	f_{\widehat{\theta}_n} ,\mathcal{S}_n\right)\right|\\
	&\leq\left|\mathcal{L}\left(f_{{\theta}^\ast}
		\right)-\mathcal{L}_\text{emp}\left(
		f_{{\theta}^\ast}
		,\mathcal{S}_n\right)\right|+2\varepsilon\\
	&\leq\max_{\theta\in \mathcal{F}_\Omega^{(\varepsilon)}}\Big|\mathcal{L}\left(  f_{{\theta}}
	\right)-\mathcal{L}_\text{emp}\left(f_{{\theta}}
,\mathcal{S}_n\right)\Big|+2\varepsilon\\
	&=\max_{\theta\in \mathcal{F}_\Omega^{(\varepsilon)}}\mathcal{E}_{\text{gap}}(  f_{{\theta}}
	,\mathcal{S}_n)+2\varepsilon,
	\end{align}
	which allows us to write:
	\begin{align}
	\mathbb{P}\Big(\left|\mathcal{E}_{\text{gen-err}}(\mathcal{S}_n)\right|>\alpha_\delta\Big)&\leq\mathbb{P}\Big(\max_{\theta\in \mathcal{F}_\Omega^{(\varepsilon)} } \mathcal{E}_{\text{gap}}(
	f_{{\theta}}
	,\mathcal{S}_n)>\alpha_\delta-2\varepsilon\Big)\\
	&\leq\sum_{\theta\in  \mathcal{F}_\Omega^{(\varepsilon)}
 }\mathbb{P}\left(\mathcal{E}_{\text{gap}}(	
f_{{\theta}}
	,\mathcal{S}_n)>\alpha_\delta-2\varepsilon\right)\\
	&\leq\left| \mathcal{F}_\Omega^{(\varepsilon)}  \right| \max_{\theta\in \mathcal{F}_\Omega^{(\varepsilon)} } \mathbb{P}\left(\mathcal{E}_{\text{gap}}(
	f_{{\theta}},\mathcal{S}_n)>\alpha_\delta-2\varepsilon\right)\\
	&\leq\left|  \mathcal{F}_\Omega^{(\varepsilon)} \right|\delta. 
	\end{align}
	Finally, $\alpha_{\delta/|   \mathcal{F}_\Omega^{(\varepsilon)} |   } $ is an upper bound with probability at least $1-\delta$ over the generalization error.
	\begin{flushright}\vspace{-0.4cm}\fbox{\phantom{\rule{.1ex}{.1ex}}}\end{flushright}
\end{proof}


\subsection{Decoupling the Decoder's Influence}

As a first step  we will decouple from the error gap the effect of the decoder. The following lemma allow us to achieve this:

\begin{lemma}\label{lem:paso2dec}
Under Assumptions \ref{asumption1}, the error gap can be bounded as
	\begin{equation}
	\mathcal{E}_{\textrm{gap}}\big(f_{{\theta}},\mathcal{S}_n\big)\leq\mathcal{E}_{\textrm{gap}}\big(q^\theta_{U|X},Q^{\theta}_{Y|U},\mathcal{S}_n)+d(\mathcal{S}_n\big),\label{eq:enc-dec}
	\end{equation}
	where $Q^{\theta}_{Y|U}(y|u)$ is the decoder induced by the encoder given in (\ref{eq:optimal_decoder}) and which we rewrite as:
	\begin{equation}
	    Q^{\theta}_{Y|U}(y|u)=P_Y(y)\frac{\mathbb{E}_{p_{X|Y}}\left[q_{U|X}^\theta(u|X)|Y=y\right]}{\mathbb{E}_{p_X}\left[q_{U|X}^\theta(u|X)\right]}
	    \label{eq:decoder_optimum}
	\end{equation}
	and where $	d(\mathcal{S}_n)$ is:
	\begin{equation}
	d(\mathcal{S}_n)=\Big|\frac{1}{n}\sum_{i=1}^nT^\theta(x_i,y_i)-\mathbb{E}_{q^{\theta}_U} \left[\textrm{KL}\left(Q^{\theta}_{Y|U}\|Q^\theta_{\widehat{Y}|U}\right) \right]\Big|
	\end{equation}
	with $T^\theta(x,y)\coloneqq\mathbb{E}_{q_{U|X}}\left[\left.\log \frac{Q^{\theta}_{Y|U}(y|U)}{Q^\theta_{\widehat{Y}|U}(y|U)} \right|X=x\right]$.
\end{lemma}
\begin{proof}
It is easy to see that:
\begin{equation}
    \ell_\theta(x,y)=\tilde{\ell}_\theta(x,y)+T^\theta(x,y),
    \label{eq:loss_decomp}
\end{equation}
where, as we already know, $\ell_\theta(x,y)$ is cross-entropy loss function (depending on the specific choice of the encoder and decoder) and $\tilde{\ell}_\theta(x,y)$ is modified cross-entropy function given by:
\[
\tilde{\ell}_\theta(x,y)\equiv 	\mathbb{E}_{q_{U|X}^\theta}\left[\left.-\log {Q}^\theta_{Y|U}(y|U)\right|X=x\right].
\]
Notice that the main difference between these two cross-entropies is that, while $\ell_\theta(x,y)$ depends on both the encoder and decoder chosen from the parametric class, $\tilde{\ell}_\theta(x,y)$ depends only on the encoder. In this case the decoder is given by (\ref{eq:decoder_optimum}), that is the optimum decoder implied by the information bottleneck criterion as discusses in Section \ref{sec:otros}. Taking expectation in (\ref{eq:loss_decomp}) with respect to $p_{XY}$ and the empirical distribution given by the set of samples $\mathcal{S}_n$ we obtain respectively:
\[\mathcal{L}(f_\theta)=\mathbb{E}_{p_{XY}}\left[\tilde{\ell}_\theta(X,Y)\right]+\mathbb{E}_{q^{\theta}_U}\left[\textrm{KL}\left(Q^{\theta}_{Y|U}\|Q^\theta_{\widehat{Y}|U}\right)\right]\]
and
\[\mathcal{L}_{emp}(f_\theta,\mathcal{S}_n)=\frac{1}{n}\sum_{i=1}^n\tilde{\ell}_{\theta}(x_i,y_i)+\frac{1}{n}\sum_{i=1}^nT^\theta(x_i,y_i),\]
where we have used that $\mathbb{E}_{p_{XY}}\left[T^\theta(X,Y)\right]=\mathbb{E}_{q^{\theta}_U}\left[ \textrm{KL}\left(Q^{\theta}_{Y|U}\|Q^\theta_{\widehat{Y}|U}\right)\right]$. Subtracting these two equations and taking absolute value at both sides we obtain:
\[\mathcal{E}_{\textrm{gap}}\big(f_{{\theta}},\mathcal{S}_n\big)=\left|\mathbb{E}_{p_{XY}}\left[\tilde{\ell}_\theta(X,Y)\right]-\frac{1}{n}\sum_{i=1}^n \big[\tilde{\ell}_{\theta}(x_i,y_i)+T^\theta(x_i,y_i)\big]+\mathbb{E}_{q^{\theta}_U}\left[\textrm{KL}\left(Q^{\theta}_{Y|U}\|Q^\theta_{\widehat{Y}|U}\right)\right]\right|.\]
Taking into account that we can define the modified error gap as:
\begin{equation}
\mathcal{E}_{\textrm{gap}}\big(q_{U|X}^\theta, Q_{Y|U}^\theta,\mathcal{S}_n\big)\equiv\left|\mathbb{E}_{p_{XY}}\left[\tilde{\ell}_\theta(X,Y)\right]-\frac{1}{n}\sum_{i=1}^n\tilde{\ell}_{\theta}(x_i,y_i)\right|
\label{eq:modified_gap}
\end{equation}
and using triangular inequality we obtain the desired result.
	\begin{flushright}\vspace{-0.4cm}\fbox{\phantom{\rule{.1ex}{.1ex}}}\end{flushright}
\end{proof}



The most important fact in this simple lemma is that the decoder influence is captured in the term $d(\mathcal{S}_n)$ (which is a simple deviation) and the encoder influence is captured through the definition of the modified loss function $\tilde{\ell}_\theta(x,y)$ and the corresponding gap $\mathcal{E}_{\textrm{gap}}(q^\theta_{U|X},Q^{\theta}_{Y|U},\mathcal{S}_n)$. It is important to note the role of the optimal decoder $Q^{\theta}_{Y|U}$ matched to the encoder $q_{U|X}^{\theta}$, on both the gap $\mathcal{E}_{\textrm{gap}}(q^\theta_{U|X},Q^{\theta}_{Y|U},\mathcal{S}_n)$ and the term $d(\mathcal{S}_n)$, where the distance between the parametric decoder $Q^\theta_{\widehat{Y}|U}$ and $Q^{\theta}_{Y|U}$ (in terms of Kullback-Leibler divergence) is explicit.

\subsection{Analysis of the Term $\mathcal{E}_{\textrm{gap}}$ }

Now we can focus on bound $\mathcal{E}_{\textrm{gap}}(q^\theta_{U|X},Q^{\theta}_{Y|U},\mathcal{S}_n)$ that depends only on the encoder. In order to do that we will consider the above presented discretization of the input space $\mathcal{X}$. Let us define, for each $y\in\mathcal{Y}$, a finite $\mathcal{B}_X$-measurable partition of the feature space $\mathcal{X}$ into $K$ cells $\{\mathcal{K}_k^{(y)}\}_{k=1}^K$ satisfying $\bigcup_{k=1}^K\mathcal{K}_k^{(y)}\equiv \mathcal{X},\;\mathcal{K}_{i}^{(y)}\cap\mathcal{K}_{j}^{(y)}=\emptyset,$ $\, \forall1\leq i<j\leq K,\;\int_{\mathcal{K}_k^{(y)}}dx>0\; \, \forall 1\leq k\leq K$. In addition,  let  $\{x^{(k,y)}\}_{k=1}^K$ denote the respective centroids for each $y\in\mathcal{Y}$, so the partition family $\mathcal{K}$ is defined as \eqref{eq:partitionfamily}.  This partition induces the probability distributions $P_{XY}^D$ \eqref{eq:PxyD}, where its marginal pmfs are $P_Y$ (true value) and a $P_X^D(x)=\sum_{y\in\mathcal{Y}}P^D_{XY}(x,y)$, which has the elements of the set $\mathcal{A}=\{x^{(k,y)}:\;1\leq k\leq K,\;y\in\mathcal{Y}\}$ as atoms. The distribution $P_{XY}^D$ and the encoder $q_{U|X}^\theta$ define a probability measure from which  $Q_{Y|U}^{D,\theta}$ in \eqref{eq:Q_Y|U^D} is obtained. Also $q_U^{D,\theta}$ is given by:
	\begin{equation}\label{eq:qUDdef}
	q_U^{D,\theta}(u) \coloneqq\sum_{k=1}^K\sum_{y\in\mathcal{Y}}q^\theta_{U|X}\big(u|x^{(k,y)}\big) P_{X|Y}(\mathcal{K}_k^{(y)}|y)P_Y(y),
	\end{equation}	
Discretization procedure introduces the magnitudes $\epsilon(\mathcal{K})$ and $r(\mathcal{K})$ defined in \eqref{eq:epsilon_r_definitions}. From the above definitions, it is implicitly understood that the partition of $\mathcal{X}$ is independent on the parametric class of the modified loss functions. However, $\epsilon(\mathcal{K})$ and $r(\mathcal{K})$ depends on the parametric class of encoders and true input distribution respectively. We are ready to establish the following lemma:
	

\begin{lemma}\label{lem:paso1disc}
Under the set of Assumptions \ref{asumption1}, let a partition family $\mathcal{K}$. The modified error gap \eqref{eq:modified_gap} can be bounded as:
\[\mathcal{E}_{\textrm{gap}}(q^\theta_{U|X},Q^{\theta}_{Y|U},\mathcal{S}_n)
\leq2\epsilon(\mathcal{K})+\mathcal{E}_{\textrm{gap}}^D(q^\theta_{U|X},Q^{D,\theta}_{Y|U},\mathcal{S}_n)\]
where $\epsilon(\mathcal{K})$ was defined in \eqref{eq:epsilon_r_definitions}, $Q^{D,\theta}_{Y|U}$ in \eqref{eq:Q_Y|U^D} and $\mathcal{E}_{\textrm{gap}}^D(q^\theta_{U|X},Q^{D,\theta}_{Y|U},\mathcal{S}_n)$ is defined as
	\begin{equation}
	\mathcal{E}_{\textrm{gap}}^D(q^\theta_{U|X},Q^{D,\theta}_{Y|U},\mathcal{S}_n)=\left|{\mathcal{L}}^D(q^\theta_{U|X},Q^{D,\theta}_{Y|U})-{\mathcal{L}}_{\textrm{emp}}^D(q^\theta_{U|X},Q^{D,\theta}_{Y|U},\mathcal{S}_n)\right|
	\end{equation}
	where
	\begin{align}
	{\mathcal{L}}^D(q^\theta_{U|X},Q^{D,\theta}_{Y|U})&=\sum_{k=1}^K\sum_{y\in\mathcal{Y}}P^D_{XY}(k,y){\ell}^D_\theta(x^{(k,y)},y),\label{eq:ld}\\
	{\mathcal{L}}_{\textrm{emp}}^D(q^\theta_{U|X},Q^{\theta}_{Y|U},\mathcal{S}_n)&=\frac{1}{n}\sum_{k=1}^{K}\sum_{\substack{i\in[1:n]\\x_i\in\mathcal{K}_k}}{\ell}^D_\theta(x^{(k,y_i)},y_i),\label{eq:ldemp}
	\end{align} 
$P^D_{XY}$ is given in \eqref{eq:PxyD} and $\ell^D_\theta$ in \eqref{eq:ellD}. 
\end{lemma}

\begin{proof}: Triangle inequality allow us to write,
	\begin{align}
\mathcal{E}_{\textrm{gap}}(q^\theta_{U|X},Q^{\theta}_{Y|U},\mathcal{S}_n) & \leq\left|{\mathcal{L}}^D(q^\theta_{U|X},Q^{D,\theta}_{Y|U})-{\mathcal{L}}_{\textrm{emp}}^D(q^\theta_{U|X},Q^{D,\theta}_{Y|U},\mathcal{S}_n))\right|\nonumber\\
&\; +\left|{\mathcal{L}}(q^\theta_{U|X},Q^{\theta}_{Y|U})-{\mathcal{L}}^D(q^\theta_{U|X},Q^{D,\theta}_{Y|U})\right|\nonumber\\
	&\; +\left|{\mathcal{L}}_{\textrm{emp}}^D(q^\theta_{U|X},Q^{D,\theta}_{Y|U},\mathcal{S}_n)-{\mathcal{L}}_{\textrm{emp}}(q^\theta_{U|X},Q^{\theta}_{Y|U},\mathcal{S}_n)\right|,\label{eq:discretizandoelgap}
	\end{align}
	where the first term is $\mathcal{E}_{\textrm{gap}}^D(q^\theta_{U|X},Q^{\theta}_{Y|U},\mathcal{S}_n)$. The other terms in \eqref{eq:discretizandoelgap} can be bounded using the fact that $P_X^D(x^{(k,y)})=\mathbb{P}\big(X\in\mathcal{K}_k^{(Y)},Y=y\big)$ and definition of $\epsilon(\mathcal{K})$ \eqref{eq:epsilon_r_definitions}. For example, for the second term we can write
\begin{align}
& \left|{\mathcal{L}}(q^\theta_{U|X},Q^{\theta}_{Y|U})-{\mathcal{L}}^D(q^\theta_{U|X},Q^{D,\theta}_{Y|U})\right|\nonumber\\
   &=\Big|\sum_{k=1}^K\sum_{y=1}^{|\mathcal{Y}|}P^D_{XY}(x^{(k,y)},y)\left(\mathbb{E}_{p_{XY}}\left[\tilde{\ell}_\theta(X,Y)|Y=y,X\in\mathcal{K}_k^{(Y)}\right]-\ell^D_\theta(x^{(k,y)},y)\right)\Big|\\
   &\leq \epsilon(\mathcal{K}).
\end{align}
The third term is treated similarly to show that:
\begin{equation}
\left|{\mathcal{L}}_{\textrm{emp}}^D(q^\theta_{U|X},Q^{\theta}_{Y|U},\mathcal{S}_n)-{\mathcal{L}}_{\textrm{emp}}(q^\theta_{U|X},Q^{\theta}_{Y|U},\mathcal{S}_n)\right|\leq \epsilon(\mathcal{K}).
\end{equation}
	\begin{flushright}\fbox{\phantom{\rule{.1ex}{.1ex}}}\end{flushright}
\end{proof}
In order to progress in the proof of the theorem, we introduce the following notation for the empirical distributions $\widehat{P}^D_{XY},\widehat{P}^D_{X}, \widehat{P}_Y, \widehat{P}^D_{X|Y}$ as the occurrence rate of $\mathcal{S}_n$; e.g. $\widehat{P}^D_{XY}(k,y)=\frac{|\left\{(x_i,y_i)\in\mathcal{S}_n:\; y_i=y,\;x_i\in\mathcal{K}_k^{(y)}\right\}|}{n}$. Also, we define distributions $\widehat{q}^{D,\theta}_U, \widehat{Q}^{D,\theta}_{Y|U}, \widehat{q}^{D,\theta}_{U|Y}$ induced from the encoder $q^\theta_{U|X}$ and the empirical distribution $\widehat{P}^D_{XY}$. In addition, notation used in the main body of paper for information magnitudes, such as entropy or mutual information, is insufficient since it does not specifies clearly the distribution with which the random variables are sampled. For this reason we introduce a change in the notation for such quantities as $\mathcal{H}(Y|U^\theta)\equiv\mathcal{H}\big({Q}^{\theta}_{Y|U}|{q}^{\theta}_U\big)$ and $\mathcal{I}(U^\theta_{(X)};X)\equiv\mathcal{I}(p_X;q^\theta_{U|X})$.

\begin{lemma}\label{lem:paso3cotaas}
Under Assumptions \ref{asumption1}, the gap $\mathcal{E}_{\textrm{gap}}^D(q^\theta_{U|X},Q^{D,\theta}_{Y|U},\mathcal{S}_n)$ can be bounded as, 
	\begin{align}
	\mathcal{E}_{\textrm{gap}}^D (q^\theta_{U|X},Q^{D,\theta}_{Y|U},\mathcal{S}_n)& \leq {\textrm{KL}}\left(\widehat{P}_{XY}^D\big \|P_{XY}^D\right)+\int_{\mathcal{U}}\phi\left(\left\|\mathbf{P}_{X}^D\!-\!\mathbf{\widehat{P}}_{X}^D\right\|_2\!\sqrt{\mathbb{V}\left(\mathbf{q}^\theta_{U|X}(u|\cdot) \right)}\right)du\nonumber\\
	&+\log \left(\frac{(4\pi eS)^{d_u/2}}{P_{Y}(y_{\min})}\right)\sqrt{|\mathcal{Y}|}\left\|\mathbf{P}_{Y}\!-\!\mathbf{\widehat{P}}_{Y}\right\|_2+\!\mathcal{O}\left(\|\mathbf{P}_{Y}\!-\!\mathbf{\widehat{P}}_{Y}\|_2^2\right)\!\nonumber\\
	&+\mathbb{E}_{P_Y}\Big[ \int_\mathcal{U}\!\phi\left(\left\|\mathbf{P}_{X|Y}^D(\cdot|Y)\!-\!\mathbf{\widehat{P}}_{X|Y}^D(\cdot|Y)\right\|_2  \!\sqrt{\mathbb{V}\left(\mathbf{q}^\theta_{U|X}(u|\cdot)\right)}\right)du\Big]\label{eq:todoslosterminos}
	\end{align}
	where $\phi(\cdot)$ is defined as
	\begin{equation}\label{eq:cotaphi}
	\phi(x)=\left\{\begin{array}{cc}0&x\leq0\\-x\log(x)&0<x<e^{-1}\\e^{-1}&x\geq e^{-1}
	\end{array}\right.
	\end{equation}
and $\mathbb{V}(\cdot)$ is defined as
	\begin{equation}
	\mathbb{V}(\mathbf{c})\coloneqq \left\|\mathbf{c}-\bar{c}\mathds{1}_a\right\|_2^2,
	\label{eq:emp_var}
	\end{equation}
	with $\mathbf{c}\in\mathbb{R}^a$, $a\in\mathbb{N}$, $\bar{c}=\frac{1}{a}\sum_{i=1}^a c_i$, and $\mathds{1}_a$ is the vector of ones of length $a$.
\end{lemma}

\begin{rem}
    Notation $\mathbf{P}_Y$ is used to think the pmf $P_Y$ as a vector $\mathbf{P}_Y=[P_Y(1),\cdots,P_Y(|\mathcal{Y}|)]$, so we can apply to it euclidean norms $\|\cdot\|_2$ and $\mathbb{V}(\cdot)$ operator. Notice, that this is well-defined as soon as the support of the considered pmfs is finite and discrete. 
\end{rem}

\begin{proof}
	Adding and subtracting $\widehat{P}_{XY}^D(x^{(k,y)},y)\mathbb{E}_{q^\theta_{U|X}}\left[\left.\log\left(\frac{1}{\widehat{Q}^{D,\theta}_{Y|U}(y|U)}\right)\right|X=x^{(k,y)}\right]$ we can write using the triangle inequality:
	\begin{align}
	& \mathcal{E}_{\textrm{gap}}^D(q^\theta_{U|X},Q^{D,\theta}_{Y|U},\mathcal{S}_n) \nonumber \\
	&=\left|\sum_{\forall (k,y)}\left[P_{XY}^D(x^{(k,y)},y)-\widehat{P}_{XY}^D(x^{(k,y)},y)\right]\mathbb{E}_{q^\theta_{U|X}}\left[\left.\log\left(\frac{1}{Q^{D,\theta}_{Y|U}(y|U)}\right)\right|X=x^{(k,y)}\right]\right|\nonumber\\
	&\leq\left| \mathcal{H}\big(Q^{D,\theta}_{Y|U}|q^{D,\theta}_U\big) - \mathcal{H}\big(\widehat{Q}^{D,\theta}_{Y|U}|\widehat{q}^{D,\theta}_U\big)\right|+ \mathbb{E}_{\widehat{q}^{D,\theta}_U}\left[ \textrm{KL}\left(\widehat{Q}^{D,\theta}_{Y|U}\big\|Q^{D,\theta}_{Y|U} \right)\right].\label{eq:segundacota}
	\end{align}
	We can bound the second term in \eqref{eq:segundacota} using the inequality: 
	\begin{equation}
	\mathbb{E}_{\widehat{q}^{D,\theta}_U}\left[\textrm{KL}\left(\widehat{Q}^{D,\theta}_{Y|U}\big\|Q^{D,\theta}_{Y|U} \right) \right]\leq\textrm{KL}\left(\widehat{Q}^{D,\theta}_{Y|U}\widehat{q}^{D,\theta}_U\big\|Q^{D,\theta}_{Y|U}q^{D,\theta}_U\right)\leq
	\textrm{KL}\left(\widehat{P}_{XY}^D\big\|P_{XY}^D\right).
	\end{equation}
	The first term of \eqref{eq:segundacota} can be bounded as:
	\begin{align}
	\left| \mathcal{H}\big(Q^{D,\theta}_{Y|U}|q^{D,\theta}_U\big ) - \mathcal{H}\big (\widehat{Q}^{D,\theta}_{Y|U}|\widehat{q}^{D,\theta}_U\big )\right| &\leq\left| \mathcal{H}(P_Y)- \mathcal{H}(\widehat{P}_{Y})\right|+\left| \mathcal{H}_d(q^D_{U}) - \mathcal{H}_d\big (\widehat{q}^{D,\theta}_{U} \big)\right|\nonumber\\
	& +\left| \mathcal{H}_d\big (q^{D,\theta}_{U|Y}|P_Y \big)- \mathcal{H}_d\big (\widehat{q}^{D,\theta}_{U|Y}|\widehat{P}_Y\big )\right|,
	\end{align}
	where $\mathcal{H}_d$ is the differential entropy. The terms $\left| \mathcal{H}_d\big(q^{D,\theta}_{U}\big )- \mathcal{H}_d\big (\widehat{q}^{D,\theta}_{U}\big )\right|$ and 
	$\big| \mathcal{H}_d\big(q^{D,\theta}_{U|Y}|P_Y\big) $ $- \mathcal{H}_d\big(\widehat{q}^{D,\theta}_{U|Y}|\widehat{P}_Y\big)\big|$ can be bounded by Lemmas~\ref{lem:Hu} and~\ref{lem:Hu|y} respectively. Finally, it is clear that
	$P_Y\mapsto  \mathcal{H}(P_Y)$ is differentiable and a first order Taylor expansion yields:
	\begin{equation}
	\mathcal{H}(P_Y)- \mathcal{H}(\widehat{P}_Y) =\left\langle\frac{\partial  \mathcal{H}(P_Y)}{\partial \mathbf{P}_{Y}},\mathbf{P}_{Y}\!-\!\mathbf{\widehat{P}}_{Y}\!\!\right\rangle+\mathcal{O}\left(\|\mathbf{P}_{Y}\!-\!\mathbf{\widehat{P}}_{Y}\|_2^2\right),
	\end{equation}
	where
	$\frac{\partial  \mathcal{H}(P_Y)}{\partial P_{Y}(y)}=-\log P_Y(y)-1$ for each $y\in\mathcal{Y}$. Then, applying Cauchy-Schwartz inequality the lemma is proved: 
	\begin{align}
	\left| \mathcal{H}(P_Y) - \mathcal{H}(\widehat{P}_Y)\right| &\leq\left|\left\langle  \log \mathbf{P}_Y,\mathbf{P}_{Y} -\mathbf{\widehat{P}}_{Y}\right\rangle\right| +  \mathcal{O}\left( \mathbf{P}_{Y} - \mathbf{\widehat{P}}_{Y}\|_2^2 \right)\\
	&\leq\left\| \log \mathbf{P}_Y\right\|_2 \Big\|\mathbf{P}_{Y}- \mathbf{\widehat{P}}_{Y}\Big \|_2 + \mathcal{O}\left(\|\mathbf{P}_{Y}\!-\!\mathbf{\widehat{P}}_{Y}\|_2^2\right)\\
	&\leq \log\left(\frac1{P_{Y}(y_{\min})}\right)\sqrt{|\mathcal{Y}|}\left\|\mathbf{P}_{Y}\!-\!\mathbf{\widehat{P}}_{Y}\right\|_2\!+\!\mathcal{O}\left(\|\mathbf{P}_{Y}\!-\!\mathbf{\widehat{P}}_{Y}\|_2^2\right).
	\end{align}
		\begin{flushright}\fbox{\phantom{\rule{.1ex}{.1ex}}}\end{flushright}
\end{proof}
The combination of lemmas \ref{lem:paso1disc}, \ref{lem:paso2dec} and \ref{lem:paso3cotaas} allow us to bound the error gap.

\subsection{Bounds Related to Concentration Inequalities}

Many terms in the above lemmas are deviation of random variables with respect to their means. As such they can be analyzed with well-known concentration inequalities. This is the case for $\textrm{KL}\big(\widehat{P}_{XY}^D\|P_{XY}^D\big)$, $\|\mathbf{P}_{X}^D-\mathbf{\widehat{P}}_{X}^D\|_2$, $\|\mathbf{P}_{Y}-\mathbf{\widehat{P}}_{Y}\|_2$, $\| \mathbf{P}_{X|Y}^D(\cdot|y)-\mathbf{\widehat{P}}_{X|Y}^D(\cdot|y)\|_2$ for $y\in\mathcal{Y}$ and $d(\mathcal{S}_n)$ simultaneously. 
With probability at least $1-\delta$, we apply Lemmas \ref{lem:diverg}, \ref{lem:pvector} and Chebyshev inequality \citep[Theorem A.16]{Devroye97a} to obtain:
\begin{align}
\textrm{KL}\left(\widehat{P}_{XY}^D\|P_{XY}^D\right) & \leq |\mathcal{X}||\mathcal{Y}|\frac{\log(n+1)}{n}+\frac{1}{n}\log\left(\frac{|\mathcal{Y}|+4}{\delta}\right)=\mathcal{O}\left(\frac{\log(n)}{n}\right), \label{eq:concentracionkl}\\
\max&\Big\{\big\|\mathbf{P}_{Y}-\mathbf{\widehat{P}}_{Y}\big\|_2,\big\|\mathbf{P}_{X}^D-\mathbf{\widehat{P}}_{X}^D\big\|_2,\big\|\mathbf{P}_{X|Y}^D(\cdot|y)-\mathbf{\widehat{P}}^D_{X|Y}(\cdot|y)\big\|_2\Big\}\nonumber\\ & \leq\frac{1+\sqrt{\log\left(\frac{|\mathcal{Y}|+4}{\delta}\right)}}{\sqrt{n}}\equiv\frac{\mathrm{B}_\delta}{\sqrt{n}},\label{eq:concentracionpv}\\
d(\mathcal{S}_n)& \leq\sqrt{\frac{|\mathcal{Y}|+4}{n\delta}}\sqrt{\text{Var}_{p_{XY}}\left(T^\theta(X,Y)\right)}.\label{eq:concentracionch}
\end{align}

Using concentration inequalities \eqref{eq:concentracionkl}, \eqref{eq:concentracionpv} and \eqref{eq:concentracionch} we have the following lemma for the error gap:

\begin{lemma}\label{lem:coninfodiscreta}
Under Assumptions \ref{asumption1},	for every $\delta\in(0,1)$ the gap satisfies:
	\begin{align}
\mathbb{P}\bigg(	\mathcal{E}_{\textrm{gap}}(f_\theta,\mathcal{S}_n)&\leq\inf_{\substack{\mathcal{K}\\\beta>0}}2\epsilon(\mathcal{K})+  A_\delta\sqrt{\mathcal{I}(P^D_X;q^\theta_{U|X})}\cdot\frac{\log(n)}{\sqrt{n}}r(\mathcal{K})\nonumber \\
& +\frac{2e^{-1}g^{D,\theta}(\beta)(1+\beta)}{\beta\sqrt{n}}\left[\sqrt{2}r(\mathcal{K})\mathrm{B}_\delta\sqrt{\mathcal{I}(P^D_X;q^\theta_{U|X})}\right]^{\frac{1}{1+\beta}}\nonumber\\
	&+\frac{\mathrm{C}_\delta+\mathrm{D}_\delta\cdot \sqrt{\mathbb{E}_{p_{XY}}\left[T^\theta(X,Y)^2\right]}}{\sqrt{n}}+\mathcal{O}\left(\frac{\log(n)}{n}\right)\bigg)\geq 1-\delta, 
	\end{align} 
	where $g^{D,\theta}(\beta)=\sqrt{\mathbb{E}_{q_U^{D,\theta}}\left[q_U^{D,\theta}(U)^{\frac{-2\beta}{1+\beta}}\right]}$.
\end{lemma}

\begin{proof}
	Using lemmas  \ref{lem:paso2dec}, \ref{lem:paso1disc}, \ref{lem:paso3cotaas} and \ref{lem:cotaphi}, with probability at least $1-\delta$ we have:
	\begin{align}
	\mathcal{E}_{\textrm{gap}}(f_\theta,\mathcal{S}_n) &\leq2\epsilon(\mathcal{K})+ \frac{\mathrm{D}_\delta}{\sqrt{n}}\text{Var}_{p_{XY}}\left(T^\theta(X,Y)\right)+\log \left(\frac{(4\pi eS)^{d_u/2}}{P_{Y}(y_{\min})}\right)\sqrt{|\mathcal{Y}|}\frac{\mathrm{B}_\delta}{\sqrt{n}}\nonumber\\
	&\quad+2\int_{\mathcal{U}}\phi\left(\frac{\mathrm{B}_\delta}{\sqrt{n}}\sqrt{\mathbb{V}\left( \textbf{q}^\theta_{U|X}(u|\cdot ) \right)} \right)du+\mathcal{O}\left(\frac{\log(n)}{n}\right)\\
	&\leq2\epsilon(\mathcal{K})+ \frac{\mathrm{D}_\delta}{\sqrt{n}}\mathbb{E}_{p_{XY}}\left[T^\theta(X,Y)^2\right]+\log \left(\frac{(4\pi eS)^{d_u/2}}{P_{Y}(y_{\min})}\right)\sqrt{|\mathcal{Y}|}\frac{\mathrm{B}_\delta}{\sqrt{n}}\nonumber\\
	&+\frac{\log(n)}{\sqrt{n}}\mathrm{B}_\delta\int_{\mathcal{U}}\sqrt{\mathbb{V}\left( \mathbf{q}^\theta_{U|X}(u|\cdot)  \right)}du \nonumber \\
 &+\frac{2(1+\beta)e^{-1}\mathrm{B}_\delta^{\frac{1}{1+\beta}}}{\beta\sqrt{n}}\int_{\mathcal{U}}\mathbb{V}\left( \mathbf{q}^\theta_{U|X}(u|\cdot)  \right)^{\frac{1}{2(1+\beta)}}du +\mathcal{O}\left(\frac{\log(n)}{n}\right)
	\end{align}
	for every $\beta>0$, where we have used that $\text{Var}_{p_{XY}}\left(T^\theta(X,Y)\right)\leq\mathbb{E}_{p_{XY}}\left[T^\theta(X,Y)^2\right]$. Next, we relate the mutual information $\mathcal{I}(P_X^D;q^\theta_{U|X})$ with  $\mathbb{V}\left( \mathbf{q}^\theta_{U|X}(u|\cdot)  \right)$. This follows from an application of Pinsker's inequality \citep[Lemma 11.6.1]{cover} $\|\mathbf{P}_1-\mathbf{P}_2\|_1^2\leq2\textrm{KL}(P_1\|P_2)$ and the fact that $\mathbb{V}(\mathbf{c}) \leq\left\|\mathbf{c}-b\mathds{1}_a\right\|_2^2\ , \ \forall b\in\mathbb{R}$:
	\begin{align}
	\mathbb{V}\left( \mathbf{q}^\theta_{U|X}(u|\cdot)  \right)&\leq\sum_{x\in\mathcal{A}}\left[q^\theta_{U|X}(u|x)-q_U^{D,\theta}(u)\right]^2\\
	&=q^{D,\theta}_U(u)^2\sum_{x\in\mathcal{A}}\left[\frac{Q^{D,\theta}_{X|U}(x|u)}{P^{D}_X(x)}-1\right]^2\\
	&\leq q^{D,\theta}_U(u)^2\left(\sum_{x\in\mathcal{A}}\left|\frac{Q^{D,\theta}_{X|U}(x|u)}{P^D_X(x)}-1\right|\right)^2\\
	&= q^{D,\theta}_U(u)^2\left(\sum_{x\in\mathcal{A}}\frac{1}{P_X^D(x)}\left|Q^{D,\theta}_{X|U}(x|u)-P_X^D(x)\right|\right)^2\\
	&\leq 2r^2(\mathcal{K}) q^{D,\theta}_U(u)^2\textrm{KL}\left(Q^{D,\theta}_{X|U}(\cdot|u)\|P^D_X\right),
	\end{align}
	where $Q^{D,\theta}_{X|U}(k|u)=\frac{q^\theta_{U|X}(u|x)P_X^D(x)}{q^{D,\theta}_U(u)}$. So, using Jensen inequality
	\begin{align}
	\int_{\mathcal{U}}\sqrt{\mathbb{V}\left( \mathbf{q}^\theta_{U|X}(u|\cdot)  \right)}du&\leq\sqrt{2}r(\mathcal{K})\mathbb{E}_{q_U^{D,\theta}}\left[\sqrt{\textrm{KL}\left(Q^{D,\theta}_{X|U}(\cdot|u)\|P^D_X\right)}\right]\\
	&\leq\sqrt{2}r(\mathcal{K})\sqrt{\mathcal{I}(P_X^D;q^\theta_{U|X})},
	\end{align}
	Similarly, we proceed using Cauchy-Swartz inequality:
	\begin{align}
	& \int_{\mathcal{U}}\mathbb{V}\left( \mathbf{q}^\theta_{U|X}(u|\cdot)  \right)^{\frac{1}{2(1+\beta)}}du  \nonumber\\
&	 \leq 2^{\frac{1}{2(1+\beta)}}r(\mathcal{K})^{\frac{1}{(1+\beta)}}\mathbb{E}_{q_U^{D,\theta}}\left[q_U^{D,\theta}(U)^{\frac{-\beta}{1+\beta}}\textrm{KL}\left(Q^{D,\theta}_{X|U}(\cdot|U)\|P^D_X\right)^{\frac{1}{2(1+\beta)}}\right]\\
	&\leq2^{\frac{1}{2(1+\beta)}}r(\mathcal{K})^{\frac{1}{(1+\beta)}}g^{D,\theta}(\beta)\sqrt{\mathbb{E}_{q_U^{D,\theta}}\left[\textrm{KL}\left(Q^{D,\theta}_{X|U}(\cdot|U)\|P^D_X\right)^{\frac{1}{(1+\beta)}}\right]}\\
	&\leq2^{\frac{1}{2(1+\beta)}}r(\mathcal{K})^{\frac{1}{(1+\beta)}}g^{D,\theta}(\beta)\cdot\mathcal{I}(P_X^D;q^\theta_{U|X})^{\frac{1}{2(1+\beta)}}.
	\end{align}	
The lemma is finally proved considering the infimum over the partition $\mathcal{K}$ and $\beta>0$.
	\begin{flushright}\vspace{-0.4cm}\fbox{\phantom{\rule{.1ex}{.1ex}}}\end{flushright}
\end{proof}

{
\begin{rem}
	It is worth mentioning the differences between our result and those presented in \citep{Shamir:2010:LGI:1808343.1808503}:  
\begin{equation}
|\mathcal{I}\big(q^{D,\theta}_U;Q^{D,\theta}_{Y|U}\big)-\mathcal{I}\big(\widehat{q}^{D,\theta}_U;\widehat{Q}^{D,\theta}_{Y|U}\big)|\leq\mathcal{O}\left(\sqrt{\mathcal{I}\big(P_X^D;q^\theta_{U|X}\big)}\frac{\log(n)}{\sqrt{n}}\right).    
\end{equation}	
	While we work with the cross-entropy gap they only bounded the mutual information gap:
	\begin{align}
	& |\mathcal{I}\big(q^{D,\theta}_U;Q^{D,\theta}_{Y|U}\big)-\mathcal{I}\big(\widehat{q}^{D,\theta}_U;\widehat{Q}^{D,\theta}_{Y|U}\big)|\nonumber \\
 &\leq\left| \mathcal{H}_d(q^{D,\theta}_{U})- \mathcal{H}_d(\widehat{q}^{D,\theta}_{U})\right|+\left| \mathcal{H}_d(q^{D,\theta}_{U|Y}|P_Y)- \mathcal{H}_d(\widehat{q}^{D,\theta}_{U|Y}|\widehat{P}_Y)\right|.
	\end{align} 
	For this reason, our proofs are substantially different. In addition, we consider continuous representations for $U$, while they work with discrete and finite alphabets. Finally, in our case, some constants were subtly reduced.
\end{rem}
}


We also have the following lemma:

\begin{lemma}
\begin{equation}
    \sqrt{\mathbb{E}_{q_U^{D,\theta}}\left[q_U^{D,\theta}(U)^{\frac{-2\beta}{1+\beta}}\right]}\leq\sup_{x,z\in\mathcal{X}}\sqrt{\int_\mathcal{U}q_{U|X}^\theta(u|x)\left(q_{U|X}^\theta(u|z)\right)^{\frac{-2\beta}{1+\beta}}du}.
\end{equation}    
\end{lemma}
\begin{proof}
Function $f(x)=x^{-\frac{2\beta}{1+\beta}}$ is convex in $x>0$ for all $\beta>0$, so we can use Jensen inequality in \eqref{eq:qUDdef}:
\begin{align}
    q_U^{D,\theta}(u)^{\frac{-2\beta}{1+\beta}}&\leq\sum_{k=1}^K\sum_{y\in\mathcal{Y}}P_{XY}(\mathcal{K}_k^{(y)},y)\left(q_{U|X}^\theta\left(u|x^{(k,y)}\right)^{\frac{-2\beta}{1+\beta}}\right).
\end{align}
With this inequality we can bound the expectation as:
\begin{align}
& \mathbb{E}_{q_U^{D,\theta}}\left[q_U^{D,\theta}(U)^{\frac{-2\beta}{1+\beta}}\right]\\
&\leq    \sum_{k=1}^K\sum_{y\in\mathcal{Y}}P_{XY}(\mathcal{K}_k^{(y)},y) \nonumber \\ 
& \times \int_\mathcal{U}\sum_{l=1}^K\sum_{y^\prime\in\mathcal{Y}}P_{XY}(\mathcal{K}_l^{(y^\prime)},y^\prime)q_{U|X}^\theta\left(u|x^{(l,y^\prime)}\right)\left(q_{U|X}^\theta\left(u|x^{(k,y)}\right)^{\frac{-2\beta}{1+\beta}}\right)du\nonumber\\
&\leq\sup_{x,z\in\mathcal{X}}\int_\mathcal{U}q_{U|X}^\theta(u|x)\left(q_{U|X}^\theta(u|z)\right)^{\frac{-2\beta}{1+\beta}}du.
\end{align}
From this last expression, the result of the lemma is inmmediate.
	\begin{flushright}\vspace{-0.4cm}\fbox{\phantom{\rule{.1ex}{.1ex}}}\end{flushright}
\end{proof}

At this point the proof of Theorem \ref{thm:regularizar} is easily concluded. In first place it is easy to see that $\mathcal{I}(p_X;q^\theta_{U|X})=\mathcal{I}(p_{XY};q^\theta_{U|X})$. As the cell $K_k^{y}$ where a particular $x\in\mathcal{X}$ belong is a deterministic function of $x$ and $y$ from the Data Processing inequality \citep[Theorem 2.8.1]{cover} $\mathcal{I}(P_X^D;q^\theta_{U|X})\leq\mathcal{I}(p_{XY};q_{U|X})=\mathcal{I}(p_{X};q^\theta_{U|X})$. The result from Lemma \ref{lem:coninfodiscreta} can be written as $\mathcal{E}_{\textrm{gap}}(q^\theta_{U|X},Q^\theta_{\widehat{Y}|U},\mathcal{S}_n)\leq\mathcal{B}(\theta,\delta)$, with probability at least $1-\delta$ and where $\mathcal{B}(\theta,\delta)$ was defined in \eqref{eq:bounderrorgap}. Using Lemma \ref{lem:coveringnet} with $\alpha_\delta=2\varepsilon+\sup_{\theta\in\mathcal{F}_{\Omega}^{(\varepsilon})}\mathcal{B}(\theta,\delta)$, the proof of Theorem \ref{thm:regularizar} is concluded.

\section{Proof of Theorem~\ref{thm:totbound}} \label{sec:proof_totbound}

Clearly, $\Omega$ is a set of functions for which we want to find conditions in order to possess the total-boundedness property under the supremum norm, i.e., 
\begin{equation}
\|f\|_{\infty}\equiv\sup_{\substack{x\in\mathcal{X}\\y\in\mathcal{Y}}}\left|f(x,y)\right|,\ \ f\in\Omega.
\end{equation}
It is well-known that the  Arzel\`{a}-Ascoli Theorem \citep[Theorem 7.25]{rudin1986principles} gives necessary and sufficient conditions for the set $\Omega$ to be totally-bounded. These conditions are the the equicontinuity and uniform boundedness of $\Omega$:
\begin{itemize}
	\item {\bf Uniform boundedness:} $\Omega$ is uniformly bounded if exists $K<\infty$ such that:
\begin{equation}
    |f(x,y)|<K, \ \ \forall (x,y)\in\mathcal{X}\times\mathcal{Y},\ \forall f\in\Omega.
\end{equation}	
	\item {\bf Equicontinuity:} $\Omega$ is equicontinuous if for every $\epsilon>0$, exists $\delta(\epsilon)>0$ such that $\forall (x_1,y_1),(x_2,y_2)\in \mathcal{X}\times\mathcal{Y}$ and $\forall f\in\Omega$:
\begin{equation}
|f(x_1,y_1)-f(x_2,y_2)|<\epsilon,\ \mbox{if}\ \ \|(x_1,y_1)-(x_2,y_2)\|<\delta(\epsilon).
\end{equation}
\end{itemize}
Notice that in our case, where $\mathcal{Y}$ is a finite and discrete space, the equicontinuity can be considered only over the input space $\mathcal{X}$, that is:
\begin{equation}
    |f(x_1,k)-f(x_2,k)|<\epsilon,\ \mbox{if}\ \ \|x_1-x_2\|<\delta(\epsilon),\ \forall k=1,\dots,|\mathcal{Y}|, 
\end{equation} 
where $\|\cdot\|$ is an appropriate norm in input space $\mathcal{X}$. As in our case $\mathcal{X}$ is contained in $\mathbb{R}^{d_x}$, this norm can be taken as the usual Euclidean norm.

The main result of this section will be to show, that under the appropriate assumptions, the set $\Omega$ in Eq. (\ref{eq:Omega_set_part}) is equicontinuous and uniformly bounded. The set of assumptions we will consider here is described below:

\begin{assumptions}\label{asumption2}: We assume the following:
	\begin{enumerate}
	\item For every $k=1,\dots, |\mathcal{Y}|$, $(w_k,b_k)\in W\subset\mathbb{R}^{|\mathcal{Y}|(d_u+1)}$ with $\mbox{diam}(W)<M_1<\infty$.
	\item For every $i=[1:d_u]$ $\alpha_i\in A\subset\mathbb{R}^l$ and $\beta_i\in B\subset\mathbb{R}^l$ with $\mbox{diam}(A)<M_2<\infty$ and $\mbox{diam}(B)<M_3<\infty$.
	\item For every $i=[1:d_u]$ functions $\mu_i(x,\beta_i)$ and $\sigma_i(x,\alpha_i)$ are uniformly bounded. That is, exists, $M_4,M_5<\infty$ such that: $|\mu_i(x,\beta_i)|<M_4$,  $|\sigma_i(x,\alpha_i)|<M_5$, $\forall x\in\mathcal{X}$, $\forall (\alpha_i,\beta_i)\in A\times B$, $i=[1:d_u]$.
	\item For every $i=1,\dots,d_u$ functions $\mu_i(x,\beta_i)$ and $\sigma_i(x,\alpha_i)$ are uniformly Lipschitz as functions of $x\in\mathcal{X}$. In precise terms: exists $K_1,K_2<\infty$ such that:
	$|\mu_i(x_1,\beta_i)-\mu_i(x_2,\beta_i)|<K_1\|x_1-x_2\|_2,\ \ \forall x_1,x_2\in\mathcal{X},\ \ \forall\beta_i\in B,\ \forall i=1,\dots, d_u$, and $|\sigma_i(x_1,\alpha_i)-\sigma_i(x_2,\alpha_i)|<K_2\|x_1-x_2\|_2,\ \ \forall x_1,x_2\in\mathcal{X},\ \ \forall\alpha_i\in A,\ \forall i=[1:d_u]$
	\item There exists $\eta>0$ such that: $\sigma_i(x,\alpha_i)>\eta,\ \forall x\in\mathcal{X},\ \ \forall\alpha_i\in A,\ \ i=[1:d_u]$.
	\end{enumerate}
\end{assumptions} 

Assumptions 1 and 2 are usually enforced in practical situations. The parameters of the encoder and decoder, found through empirical risk minimization usually belong to compact sets in the parameter space, practically avoiding that they could diverge during the training phase. In many cases, this is usually enforced through the use of a proper regularization for the empirical risk. Assumption 3 and 4 are typically satisfied for several feed-forward architectures. For example, for RELU activation functions, assumption 2 is sufficient for assumption 4 to be true. With the additional assumption that $\mathcal{X}$ is closed and bounded, assumption 3 will also be true for RELU activation functions. For sigmoid activations, assumption 3 is valid even if $\mathcal{X}$ is not bounded. Constants $K_1$ and $K_2$ can be easily written in terms of the number $L$ of layers of the feed-forward architecture used and properties of the activation functions used\footnote{In order to keep the expressions simpler we will not do this.}. Moreover, the  Assumption 4 can also be relaxed for asking only for equicontinuity for the functions $\mu_i(x,\beta_i)$ and $\sigma_i(x,\alpha_i)$ without any loss. Assumption 5 is needed to avoid degeneration of the Gaussian encoders. It is also easy to enforce a typical parameter learning scenario.

By Arzel\`{a}-Ascoli Theorem we will need to show that $\Omega$ is uniformly bounded and equicontinuous. Let us begin with the uniform boundedness property. Using 
\[\ell_{\theta_E,\theta_D}(x,k)=\mathbb{E}_{q_{U|X}^{\theta_E}}\left[\left.-\log Q_{\hat{Y}|U}^{\theta_D}(k|U)\right|X=x\right],\]
we can write:
\begin{align}
&|\ell_{\theta_E,\theta_D}(x,k)|\nonumber\\
=& \left|\int_{\mathcal{U}}\prod_{i=1}^{d_u} \mathcal{N}\left(\mu_i(x,\beta_i),\sigma_i^2(x,\alpha_i)\right)\left[\langle w_k,u\rangle+b_k-\log\left(\sum_{i=1}^{|\mathcal{Y}|}\exp\left\{\langle w_i,u\rangle+b_i\right\}\right)\right]du\right|\\
\leq & |\langle w_k\mathbb{E}_{q_{U|X}^{\theta_E}}\left[u\right]\rangle|+|b_k|+\left|\int_{\mathcal{U}}\prod_{i=1}^{d_u} \mathcal{N}\left(\mu_i(x,\beta_i),\sigma_i^2(x,\alpha_i)\right)\log\left(\sum_{i=1}^{|\mathcal{Y}|}\exp\left\{\langle w_i,u\rangle+b_i\right\}\right)du\right|\\
\leq& \|w_k\|_2\left(\sum_{i=1}^{d_u}\mu_i^2(x,\beta_i)\right)^{1/2}+|b_k|\nonumber \\
 +&\int_{\mathcal{U}}\prod_{i=1}^{d_u} \mathcal{N}\left(\mu_i(x,\beta_i),\sigma_i^2(x,\alpha_i)\right)\times\left|\log\left(\sum_{i=1}^{|\mathcal{Y}|}\exp\left\{\langle w_i,u\rangle+b_i\right\}\right)\right|du,
\end{align}
where we have used the subaddtivity of absolute value and Cauchy-Schwartz inequality. Using the underlying assumptions, it is not difficult to check that 
\begin{equation}
    \|w_k\|_2\left(\sum_{i=1}^{d_u}\mu_i^2(x,\beta_i)\right)^{1/2}+|b_k|< M_1d_u^{1/2}M_4+M_1.
\end{equation}
For the other term we can use the following easy to obtain inequality:
\begin{equation}
\left|\log\left(\sum_{i=1}^{|\mathcal{Y}|}\exp\left\{\langle w_i,u\rangle+b_i\right\}\right)\right|\leq \max_{i=[1:|\mathcal{Y}|]}\left\{\left|\langle w_i,u\rangle+b_i\right|\right\}+\log|\mathcal{Y}|.
\label{eq:bound_soft_max}
\end{equation}
Then, 
\begin{align}
&\int_{\mathcal{U}}\prod_{i=1}^{d_u} \mathcal{N}\left(\mu_i(x,\beta_i),\sigma_i^2(x,\alpha_i)\right)\left|\log\left(\sum_{i=1}^{|\mathcal{Y}|}\exp\left\{\langle w_i,u\rangle+b_i\right\}\right)\right|du\nonumber\\
&\qquad\leq \mathbb{E}_{q_{U|X}^{\theta_E}}\left[\max_{i=[1: |\mathcal{Y}|]}\left\{|\langle w_i,u\rangle+b_i|\right\}|X=x\right]+\log|\mathcal{Y}|.
\end{align}
It is straightforward to write:
\begin{align}
\mathbb{E}_{q_{U|X}^{\theta_E}}\left[\max_{i=[1: |\mathcal{Y}|]}\left\{|\langle w_i,u\rangle+b_i|\right\}\right]&\leq \mathbb{E}_{q_{U|X}^{\theta_E}}\left[\max_{i=[1: |\mathcal{Y}|]}\left\{\|w_i\|_2\|U\|_2\right\}|X=x\right]+\max_{i=[1: |\mathcal{Y}|]}|b_i|\\
&\leq M_1\cdot \mathbb{E}_{q_{U|X}^{\theta_E}}\left[\|U\|_2|X=x\right]+M_1\\
&\leq M_1\sqrt{\mathbb{E}_{q_{U|X}^{\theta_E}}\left[\|U\|_2^2|X=x\right]}+M_1\\
&= M_1\sqrt{\sum_{i=1}^{d_u}\left(\mu_i^2(x,\beta_i)+\sigma_i^2(x,\alpha_i)\right)}+M_1\\
&\leq M_1 d_u^{1/2}(M_4^2+M_5^2)^{1/2}+M_1,
\end{align}
where we have use Cauchy-Schwartz and Jensen inequality and the set to Assumptions \ref{asumption2}. Combining the above results we obtain that $\Omega$ is uniformly bounded.

For the equicontinuity of $\Omega$ we can write:
\begin{align}
& |\ell_{\theta_E,\theta_D}(x_2,k)-\ell_{\theta_E,\theta_D}(x_2,k)| \nonumber \\  
& \leq\int_{\mathcal{U}}\left|q_{U|X}^{\theta_E}(u|x_2)-q_{U|X}^{\theta_E}(u|x_1)\right|\times\left|
	\log Q_{\hat{Y}|U}^{\theta_D}(k|u)\right|du\\
& \leq \int_{\mathcal{U}}\left|q_{U|X}^{\theta_E}(u|x_2)-q_{U|X}^{\theta_E}(u|x_1)\right|\times(M_1\|u\|_2+M_1+\log|\mathcal{Y}|)du, 
\end{align}
where we have used (\ref{eq:bound_soft_max}) and the set of Assumptions \ref{asumption2}. Consider the total variation for two probability densities, which can be written as:
\begin{equation}
    \mbox{TV}(p;q)\equiv \frac{1}{2}\int|p(x)-q(x)|dx.
\end{equation}
Using this definition we can write:
\begin{align}
	|\ell_{\theta_E,\theta_D}(x_2,k)-\ell_{\theta_E,\theta_D}(x_2,k)| 
 & \leq(M_1+\log|\mathcal{Y}|)\mbox{TV}\left(q_{U|X}^{\theta_E}(\cdot |x_2);q_{U|X}^{\theta_E}(\cdot |x_1)\right)\nonumber\\
	&\qquad+M_1\int_\mathcal{U}\|u\|_2\left|q_{U|X}^{\theta_E}(u|x_2)-q_{U|X}^{\theta_E}(u|x_1)\right|du.
\end{align}
The second term in the above equation can be bounded using Cauchy-Schwartz yielding: 
\begin{align}
& \int_\mathcal{U}\|u\|_2\left|q_{U|X}^{\theta_E}(u|x_2)-q_{U|X}^{\theta_E}(u|x_1)\right|du \nonumber \\
&\leq \sqrt{2}\cdot \mbox{TV}^{1/2}\left(q_{U|X}^{\theta_E}(\cdot |x_2);q_{U|X}^{\theta_E}(\cdot |x_1)\right) \left(\int_\mathcal{U}\|u\|_2^2\left|q_{U|X}^{\theta_E}(u|x_2)-q_{U|X}^{\theta_E}(u|x_1)\right|du\right)^{1/2}\\
&\leq \sqrt{2}\cdot \mbox{TV}^{1/2}\left(q_{U|X}^{\theta_E}( \cdot |x_2);q_{U|X}^{\theta_E}(\cdot |x_1)\right) \left(\mathbb{E}_{q_{U|X}^{\theta_E}}\left[\|U\|_2^2|X=x_2\right]+\mathbb{E}_{q_{U|X}^{\theta_E}}\left[\|U\|_2^2|X=x_1\right]\right)^{1/2}\\
&\leq\sqrt{2}\cdot d_u^{1/2}\left(M_4^2+M_5^2\right)^{1/2}\mbox{TV}^{1/2}\left(q_{U|X}^{\theta_E}( \cdot |x_2);q_{U|X}^{\theta_E}(\cdot |x_1)\right).
\end{align}
We see that the equicontinuity for $\Omega$ depends on the continuity properties of the variational distance for two Gaussians encoders with inputs $x_1$ and $x_2$. The variational distance for two Gaussian pdfs is difficult to compute in close form \citep{Devroye_Mehrabian_Reddad_2020}. However, it can be easily bounded using Pinsker Inequality \citep{pinsker1964information}:
\begin{equation}
    TV(p;q)\leq\sqrt{\frac{1}{2}\textrm{KL}(p||q)}.
\end{equation}
As the encoders are Gaussian, the KL divergence can be easily computed and thus, 
\begin{align}
& \mbox{TV}\left(q_{U|X}^{\theta_E}(\cdot|x_2);q_{U|X}^{\theta_E}(\cdot |x_1)\right) \nonumber \\
& \leq \left(\frac{1}{4}\sum_{i=1}^{d_u}\frac{\sigma_i^2(x_2,\alpha_i)}{\sigma_i^2(x_1,\alpha_i)}-1+\log\frac{\sigma_i^2(x_1,\alpha_i)}{\sigma_i^2(x_2,\alpha_i)}+\frac{\left[\mu_i^2(x_2,\beta_i)-\mu_i^2(x_1,\beta_i)\right]^2}{\sigma_i^2(x_1,\alpha_i)}\right)^{1/2}.
\end{align}
It will suffice to analyze each of the following quantities:
\begin{equation}
    \frac{\sigma_i^2(x_2,\alpha_i)}{\sigma_i^2(x_1,\alpha_i)}-1+\log\frac{\sigma_i^2(x_1,\alpha_i)}{\sigma_i^2(x_2,\alpha_i)}+\frac{\left[\mu_i^2(x_2,\beta_i)-\mu_i^2(x_1,\beta_i)\right]^2}{\sigma_i^2(x_1,\alpha_i)}.
\end{equation}
Using items 3 and 5 in Assumptions \ref{asumption2} we can write:
\begin{align}
|\sigma_i^2(x_1,\alpha_i)-\sigma_i^2(x_2,\alpha_i)|&\leq|\sigma_i(x_1,\alpha_i)-\sigma_i(x_2,\alpha_i)||\sigma_i(x_1,\alpha_i)+\sigma_i(x_2,\alpha_i)|\\
&\leq 2M_5K_2\|x_2-x_1\|_2.
\end{align}
Using the fact that $\sigma_i^2(x_1,\alpha_i)\leq \sigma_i^2(x_2,\alpha_i)+2M_5K_2\|x_2-x_1\|_2$ and Claim 5 in Assumptions \ref{asumption2}, we can write:
\begin{align}
\log\frac{\sigma_i^2(x_1,\alpha_i)}{\sigma_i^2(x_2,\alpha_i)}&\leq \log\left(1+\frac{2M_5\|x_2-x_1\|_2}{\sigma_i^2(x_2,\alpha_i)}\right)\\
    &\leq \frac{2M_5K_2\|x_2-x_1\|_2}{\sigma_i^2(x_2,\alpha_i)}\\
    &\leq \frac{2M_5K_2\|x_2-x_1\|_2}{\eta^2}.
\end{align}
Similarly, we have 
\begin{align}
    \frac{\sigma_i^2(x_2,\alpha_i)}{\sigma_i^2(x_1,\alpha_i)}-1&\leq \frac{1}{\sigma_i^2(x_1,\alpha_i)}|\sigma_i^2(x_1,\alpha_i)-\sigma_i^2(x_2,\alpha_i)|\\
    &\leq\frac{2M_5K_2\|x_2-x_1\|_2}{\eta^2}.
\end{align}
Finally,
\begin{equation}
    \frac{\left[\mu_i^2(x_2,\beta_i)-\mu_i^2(x_1,\beta_i)\right]^2}{\sigma_i^2(x_1,\alpha_i)}\leq\frac{K_1^2\|x_1-x_2\|^2_2}{\eta^2}.
\end{equation}
Combining all the above results we obtain:
\begin{align}
&|\ell_{\theta_E,\theta_D}(x_2,k)-\ell_{\theta_E,\theta_D}(x_2,k)|
\leq (M_1+\log|\mathcal{Y}|)\frac{d_u^{1/2}}{2\eta}\left(K_1^2\|x_1-x_2\|_2^2+4M_5K_2\|x_1-x_2\|_2\right)^{1/2}\nonumber\\
& +\frac{M_1d_u^{3/4}(M_4^2+M_5^2)^{1/2}}{\eta^{1/2}}\left(K_1^2\|x_1-x_2\|_2^2+4M_5K_2\|x_1-x_2\|_2\right)^{1/4},
\end{align}
from which equicontinuity immediately follows.

\section{Auxiliary Results}\label{app:auxiliary-results}

In this Appendix, some auxiliary facts which are used in proof of the main result are presented.

\subsection{Some Basic Results}

\begin{lemma}\citep[Theorem 11.2.1]{cover}\label{lem:diverg}
	Let ${P}\in\mathcal{P}(\mathcal{X})$ be a discrete probability distribution and let $\widehat{P}$ be its empirical estimation over a $n$-data set. Then, 
	\begin{equation}
	\mathbb{P}\left(\textrm{KL}(\widehat{P}\|P)\leq |\mathcal{X} | \frac{\log(n+1)}{n}+\frac{1}{n}\log(1/\delta)\right)\geq1-\delta,
	\end{equation}
	for all $\delta\in(0,1)$.
\end{lemma}


\begin{lemma}[Application of McDiarmid's Inequality]\label{lem:pvector}
	Let ${P}\in\mathcal{P}(\mathcal{X})$ be any probability distribution and let $\hat{{P}}$ be its empirical estimation over a $n$-data set. Then,
	\begin{equation}
	\mathbb{P}\left(\|\mathbf{P}-\mathbf{\hat{{P}}}\|_2\leq\frac{1+\sqrt{\log(1/\delta)}}{\sqrt{n}}\right)\geq1-\delta,
	\end{equation}
	for all $\delta\in(0,1)$.
\end{lemma}

\begin{lemma}[Modified result from \citep{Shamir:2010:LGI:1808343.1808503}]\label{lem:Hu}
Let $X$ and $U$ be two random variables ($X$ is discrete and $U$ is continuous) distributed according to  $P_X$  and $q_{U|X}$, respectively  and let $\widehat{P}_X$ be its empirical estimation over a set of sample size $n$. Then, 
	\begin{equation}
	\left| \mathcal{H}_d\left(q_U\right)- \mathcal{H}_d\left(\widehat{q}_U\right)\right|\leq\int_\mathcal{U}\phi\left(\|\mathbf{P}_X-\mathbf{\widehat{P}}_X\|_2 \sqrt{\mathbb{V}\big( \mathbf{q}_{U|X}(u|\cdot) \big) }\right)du, 
	\end{equation}
	where $q_U(u)=\mathbb{E}_{P_X}\left[q_{U|X}(u|X)\right]$; $\widehat{q}_U(u)=\mathbb{E}_{\widehat{P}_X}\left[q_{U|X}(u|X)\right]$; $\phi(\cdot)$ is defined in \eqref{eq:cotaphi} and $\mathbb{V}(\cdot)$ in \eqref{eq:emp_var} for $\|\mathbf{P}^D_X-\mathbf{\widehat{P}}^D_X\|_2$  small enough\footnote{In the present context, this magnitude is $\mathcal{O}(n^{-1/2})$.}.
\end{lemma}

\subsection{Additional  Auxiliary Results}

\begin{lemma}\label{lem:cotaphi}
	Let $n\geq a^{2}e^{2}$ with $a\geq0$, then $	\phi\left(\frac{a}{\sqrt{n}}\right)\leq\frac{a}{2}\frac{\log(n)}{\sqrt{n}}+\frac{(1+\beta)e^{-1}}{\beta}\frac{a^{\frac{1}{1+\beta}}}{\sqrt{n}}$ for every $\beta>0$.
\end{lemma}
\begin{proof}
	Function $\phi(\cdot)$ is defined in \eqref{eq:cotaphi}, for  $n\geq a^{2}e^{2}$ as
	\begin{equation}
	\phi\left(\frac{a}{\sqrt{n}}\right)=\frac{a}{2}\frac{\log(n)}{\sqrt{n}}+\frac{a\log\left(\frac{1}{a}\right)}{\sqrt{n}}.
	\end{equation} 
	In order to bound the second summand, we look for the maximum of the following function:
	\begin{equation}
	f_\beta(x)=\frac{x\log\left(\frac{1}{x}\right)}{x^{\frac{1}{1+\beta}}}=-x^{\frac{\beta}{1+\beta}}\log\left(x\right).
	\end{equation}
	It is easy to check its derivative:
	\begin{align}
	f_\beta^\prime(x)&=-\frac{\beta}{1+\beta}x^{\frac{\beta}{1+\beta}-1}\log\left(x\right)-x^{\frac{\beta}{1+\beta}-1}=-x^{\frac{\beta}{1+\beta}-1}\left[\frac{\beta}{1+\beta}\log(x)+1\right].
	\end{align}
	Derivative is null in $e^{-\frac{1+\beta}{\beta}}$ and this point is a maximum because $f_\beta^\prime(x)>0$ for $x<e^{-\frac{1+\beta}{\beta}}$ and $f_\beta^\prime(x)<0$ for $x>e^{-\frac{1+\beta}{\beta}}$. Finally,
	\begin{equation}
	x\log\left(\frac{1}{x}\right)=f_\beta(x)x^{\frac{1}{1+\beta}}\leq f_\beta(e^{-\frac{1+\beta}{\beta}})x^{\frac{1}{1+\beta}}=\frac{1+\beta}{\beta}e^{-1}x^{\frac{1}{1+\beta}}.
	\end{equation}
		\begin{flushright}\fbox{\phantom{\rule{.1ex}{.1ex}}}\end{flushright}
\end{proof}

\begin{lemma}\label{lem:Hu|y} 
	Under Assumptions \ref{asumption1}:
	\begin{align}
	&\left| \mathcal{H}_d\big(q^{D,\theta}_{U|Y}|P_Y\big)- \mathcal{H}_d\big(\widehat{q}^{D,\theta}_{U|Y}|\widehat{P}_Y\big)\right|\leq \|\mathbf{P}_Y - \mathbf{\widehat{P}}_Y\|_2\frac{\sqrt{|\mathcal{Y}|}d_u}{2}\log\left(4\pi eS\right)\nonumber\\
	&\qquad+\mathbb{E}_{P_Y}\!\left[ \int_\mathcal{U}\!\phi\left(\left\|\mathbf{P}^D_{X|Y}(\cdot|Y)\!-\!\mathbf{\widehat{P}}^D_{X|Y}(\cdot|Y)\right\|_2\!\sqrt{\mathbb{V}\big( \mathbf{q}^\theta_{U|X}(u|\cdot )  \big) }\right)du\right],
	\end{align}
	for $\max_y\|\mathbf{P}^D_{X|Y}(\cdot|y)-\mathbf{\widehat{P}}^D_{X|Y}(\cdot|y)\|_2$ small enough.
\end{lemma}
\begin{proof}
	Using triangle and Cauchy-Swartz inequalities, we obtain: 	
	\begin{align}
	| \mathcal{H}_d\big (q^{D,\theta}_{U|Y}|P_Y\big )- \mathcal{H}_d\big (\widehat{q}^{D,\theta}_{U|Y}|\widehat{P}_Y\big )| 
	&=\left|\sum_{y\in\mathcal{Y}}P_Y(y)\mathcal{H}_d(q^{D,\theta}_{U|Y}(\cdot|y))-\widehat{P}_Y(y)\mathcal{H}_d(\widehat{q}^{D,\theta}_{U|Y}(\cdot|y))\right|\\
	&\leq\left|\sum_{y\in\mathcal{Y}}P_Y(y)\left(\mathcal{H}_d(q^{D,\theta}_{U|Y}(\cdot|y))-\mathcal{H}_d(\widehat{q}^{D,\theta}_{U|Y}(\cdot|y))\right)\right|\nonumber\\
	&\qquad+\left|\sum_{y\in\mathcal{Y}}\left(P_Y(y)-\widehat{P}_Y(y)\right)\mathcal{H}_d(\widehat{q}^{D,\theta}_{U|Y}(\cdot|y))\right|\\
	&\leq\sum_{y\in\mathcal{Y}}P_Y(y)\left|\mathcal{H}_d(q^{D,\theta}_{U|Y}(\cdot|y))-\mathcal{H}_d(\widehat{q}^{D,\theta}_{U|Y}(\cdot|y))\right|\nonumber\\
	&\qquad+\|\mathbf{P}_Y\!-\!\mathbf{\widehat{P}}_Y\|_2\cdot\sqrt{\sum_{y\in\mathcal{Y}}\mathcal{H}^2_d(\widehat{q}^{D,\theta}_{U|Y}(\cdot|y))} . \label{eq:ultimosdosdeqy|u}
	\end{align}
	We apply Lemma \ref{lem:Hu} to the first term in \eqref{eq:ultimosdosdeqy|u} and we obtain:
	\begin{align}
	&\sum_{y\in\mathcal{Y}}P_Y(y)\left|\mathcal{H}_d(q^{D,\theta}_{U|Y}(\cdot|y))-\mathcal{H}_d(\widehat{q}^{D,\theta}_{U|Y}(\cdot|y))\right|\nonumber\\
	&\qquad\leq\mathbb{E}_{P_Y}\!\left[ \int_\mathcal{U}\!\phi\left(\left\|\mathbf{P}^D_{X|Y}(\cdot|Y)\!-\!\mathbf{\widehat{P}}^D_{X|Y}(\cdot|Y)\right\|_2\!\sqrt{\mathbb{V}\big( \mathbf{q}^{\theta}_{U|X}(u|\cdot )  \big) }\right)du\right].
	\end{align}
	For the second term of \eqref{eq:ultimosdosdeqy|u} we bound the differential entropy using Hadamard's inequality and the fact that it is maximized for Gaussian random variables: 
	\begin{align}
	\sum_{y\in\mathcal{Y}}\mathcal{H}^2_d(\widehat{q}^{D,\theta}_{U|Y}(\cdot|y))&\leq\sum_{y\in\mathcal{Y}}\frac{1}{4}\log^2\left[(2\pi e)^{d_u}\text{det}\left(\Sigma_{\widehat{q}^{D,\theta}_{U|Y}}(U|Y=y)\right)\right]\\
	&\leq\sum_{y\in\mathcal{Y}}\frac{1}{4}\left(\sum_{j=1}^{d_u}\log\left[2\pi e\cdot\text{Var}_{\widehat{q}^{D,\theta}_{U_j|Y}}(U_j|Y=y)\right]\right)^2,
	\end{align}
	where $\mathbb{E}_{\widehat{q}^{D,\theta}_{U|Y}}\big[U|Y=y \big]$ denotes  the covariance matrix associated to  $\widehat{q}^{D,\theta}_{U|Y}(\cdot|y)$. In order to bound the last term, we make use the law of total variance:
	\begin{align}
	\text{Var}_{\widehat{q}^{D,\theta}_{U_j|Y}}(U_j|Y=y)
	&\;=\mathbb{E}_{\widehat{P}_{X|Y}^D}\left[\text{Var}_{{q}^\theta_{U_j|X}}(U_j|X)\Big|Y=y\right]+\text{Var}_{\widehat{P}_{X|Y}^D}\left(\mathbb{E}_{q^\theta_{U_j|X}}\left[U_j|X\right]\Big|Y=y\right)\\
	&\;\leq\mathbb{E}_{\widehat{P}_{X|Y}^D}\left[\text{Var}_{{q}^\theta_{U_j|X}}(U_j|X)+\mathbb{E}_{q^\theta_{U_j|X}}^2\left[U_j|X\right]\Big|Y=y\right]\\
	&\;\leq 2S,
	\end{align}
	where we use that $\text{Var}_{{q}^\theta_{U_j|X}}(U_j|X=x)\leq S$ and $\mathbb{E}_{q^\theta_{U_j|X}}\left[U_j|X=x\right]\leq \sqrt{S}$ for all $x\in\mathcal{X}$. Finally,
	\begin{align}
	\sum_{y\in\mathcal{Y}}\mathcal{H}^2_d(\widehat{q}^{D,\theta}_{U|Y}(\cdot|y))\leq\frac{|\mathcal{Y}|m^2}{4}\log^2\left(4\pi eS\right).
	\end{align}
		\begin{flushright}\fbox{\phantom{\rule{.1ex}{.1ex}}}\end{flushright}
\end{proof}

{
\section{Complementary Simulations}\label{app:experimentos}

As our main goal is not to present a new classification methodology, with competitive  state-of-the-art methods, in Section \ref{sec:experimentos} we restricted ourselves to small subsets of traditional image databases. 
In order to complement the aforementioned Section \ref{sec:experimentos} and to corroborate that the results obtained are not a consequence of the specifics of those simulations, in this Appendix we present some numerical results that complement the previous analysis. In particular present new numerical experiments with following goals in mind:
\begin{itemize}
    \item Although several cases were shown in Section 5 where the behavior of generalization error and mutual information is similar, it is desired to verify the implications of this phenomenon in the expected risk whose minimization is the actual goal in practice.
    \item To corroborate that the observed results are not a consequence of using small databases and simpler algorithms, we perform numerical experiments with full databases and more complex networks such as convolutional ones.
    \item As the reviewer suggested we include an experiment that considers an unstructured database: the Forest Cover Type Dataset \citep{arbolitosdataset}.
\end{itemize}

\subsection{Expected Risk Behaviour}
\begin{figure}[t]
\begin{subfigure}{.32\textwidth}
  \centering
  \includegraphics[width=\linewidth]{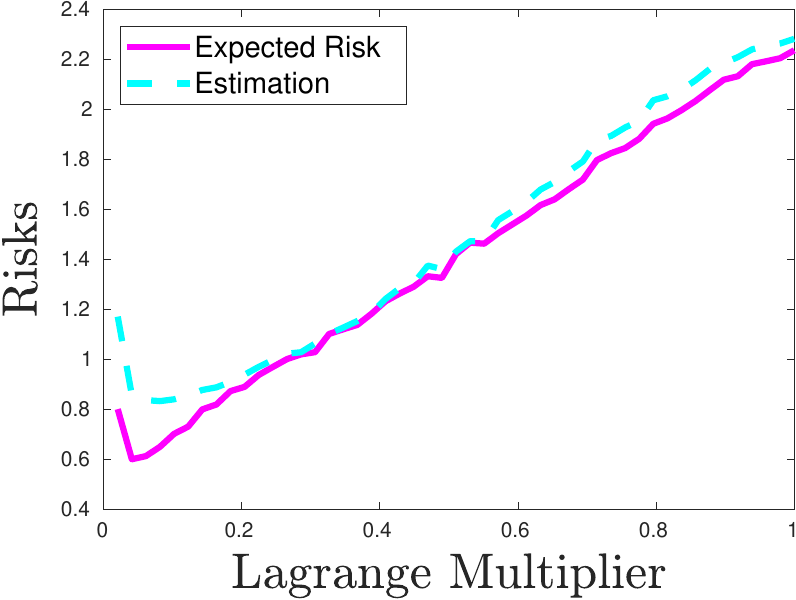}  
  \caption{Normal Encoder}
  \label{fig:app_ER_normal}
\end{subfigure}
\begin{subfigure}{.32\textwidth}
  \centering
  \includegraphics[width=\linewidth]{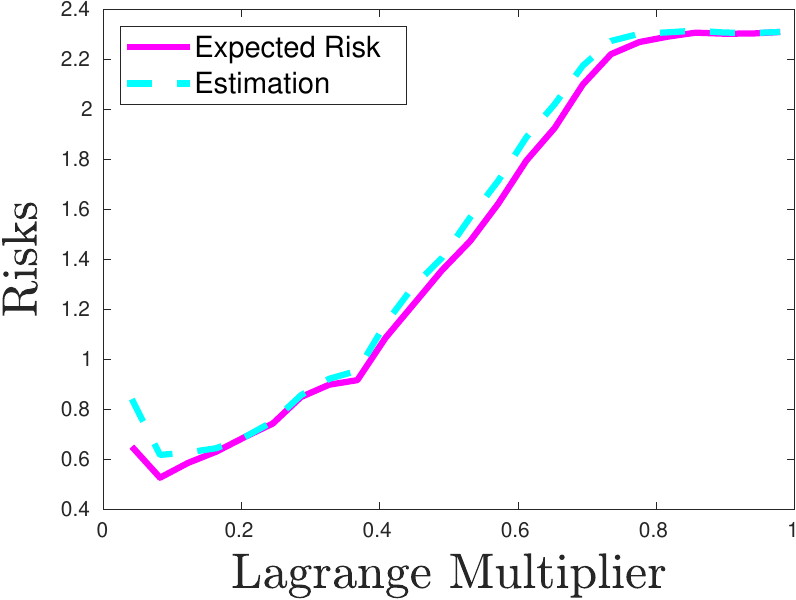}  
  \caption{LogNormal Encoder}
  \label{fig:app_ER_lognormal}
\end{subfigure}
\begin{subfigure}{.32\textwidth}
  \centering
  \includegraphics[width=\linewidth]{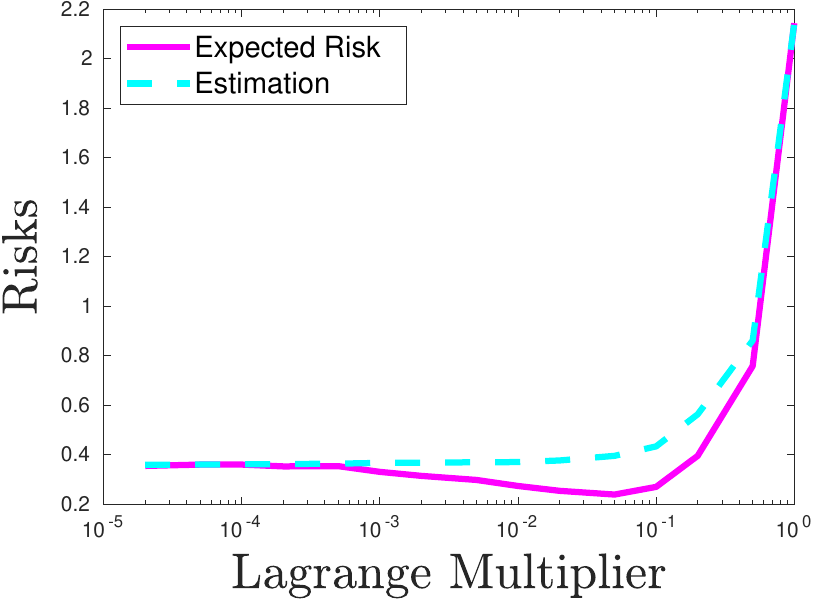}  
  \caption{RBM Encoder}
  \label{fig:app_ER_rbm}
\end{subfigure}
\caption{Expected risk and the bound \eqref{eq:app_bound} estimation behaviour for: (a) Normal Encoder, (b) LogNormal Encoder, and (c) RBM Encoder.}
\label{fig:expected_risks}
\end{figure}

In Section \ref{sec:experimentos} we show that the behaviour of the generalization error and mutual information are close in some practical examples. We considered the main term from Theo. \ref{thm:regularizar} and tried to validate:
\begin{equation}\label{eq:app_bound}
\mathcal{L}\big(f_{\widehat{\theta}_n} \big)\leq \mathcal{L}_\text{emp}\big(f_{\widehat{\theta}_n},\mathcal{S}_n\big)+a\cdot\sqrt{\mathcal{I}\big( U^{\widehat{\theta}_n}_{(X)}; X \big)}+b    
\end{equation}
for $a,b$ fixed constants and where mutual information is estimated from the training set. Unfortunately $a$ and $b$ are unknown, and as Theo. \ref{thm:regularizar} shows, depends of several quantities that are difficult to estimate. In order to check the role of $\sqrt{\mathcal{I}\big( U^{\widehat{\theta}_n}_{(X)}; X \big)}$ we will consider that those values of $a,b$ do not change when the training Lagrange multiplier, used in the numerical optimization of the encoder/decoder pair, varies. In this sense, the tightness of the bound \eqref{eq:app_bound} can be evaluated by choosing the best possible fixed constants $a,b$. 

In Fig. \ref{fig:expected_risks} it can be seen how tight the bound \eqref{eq:app_bound} can be on MNIST dataset with the setup presented in Section \ref{sec:experimentos} for Normal (Fig. \ref{fig:app_ER_normal}), Log-Normal (Fig. \ref{fig:app_ER_lognormal}) and RBM encoder(Fig. \ref{fig:app_ER_rbm}). Although there are no guarantees that this choice of values $a$ and $b$ is representative of the magnitudes presented in the Theo. \ref{thm:regularizar}, it can be seen that the bound in (\ref{eq:app_bound}) is very tight for $a,b$ fixed for the different values of Lagrange multipliers showed. 

\subsection{Mutual Information with more data and Convolutional nets}

\begin{figure}
    \centering
    \includegraphics[width=0.8\textwidth]{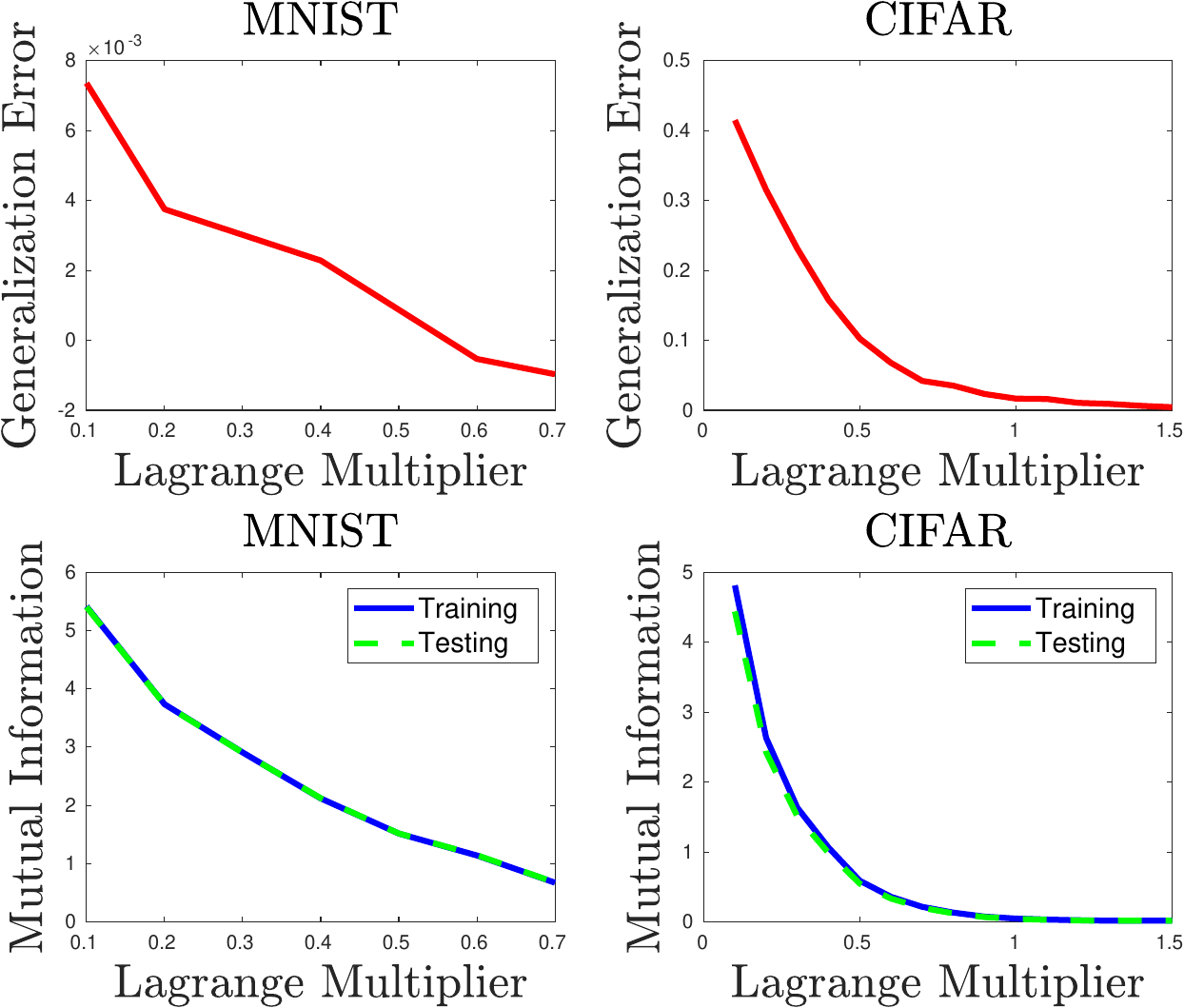}
    \caption{Generalization error and mutual information for a Convolutional Normal encoder architecture. Curves on the left correspond to experiments on the MNIST database and those on the right correspond to CIFAR-10.}
    \label{fig:full}
\end{figure}

As our main goal is not to present a new classification methodology, with competitive  state-of-the-art methods, in Section \ref{sec:experimentos} we restricted ourselves to small subsets of databases, as motivated by \citep{Neyshabur2017GeometryOO}. This data reduction suggests the use of small architectures, especially considering that variational methods usually have fewer parameters than classical ones. The question then arises whether the conclusions generated in the aforementioned section can be extended to the case where the entire database and more complex architectures such as convolutional neural networks are used. In this section we study the behaviour of generalization error and mutual information for convolutional variational algorithms using full MNIST and CIFAR-10 datasets. 

Different architecture decisions were made for each database. For MNIST dataset we considered a Normal encoder (see Section \ref{sec:normal_encoder}) composed of two layers of kernel size $3$, ReLU activation and padding ``valid'' with $32$ and $64$ filters respectively. Each of these layers are followed with a max-pooling layer of pool size $2\times2$. These layers (flatten) are followed by another linear one for each parameter ($\mu$ and $\log\sigma^2$) with $1024$ hidden units, that is each parameter, $\mu$ and $\log\sigma^2$, are a four-layers network where the first three are common to both. We chose a learning rate of $0.001$ (ADAM optimizer), a batch-size of $32$ and we trained during $50$ epochs. We repeated simulations $3$ times for each Lagrange multiplier value and report the average metrics. We considered Lagrange multipliers where the test accuracy exceeds $90\%$, exceeding $99\%$ for some values.

For CIFAR-10 dataset we considered a Normal encoder composed of three layers of kernel size $4$, strides $2$, ReLU activation and padding ``same'' with $64$, $128$ and $512$ filters respectively. These layers (flatten) are followed by another linear one for each parameter ($\mu$ and $\log\sigma^2$) with $1024$ hidden units, that is each parameter, $\mu$ and $\log\sigma^2$, is a four-layers network where the first three are common to both. We chose a learning rate of $0.0005$ (ADAM optimizer), a batch-size of $250$ and we trained during $50$ epochs. We repeated simulations $3$ times for each Lagrange multiplier value and report the average metrics. We considered Lagrange multipliers where the test accuracy exceeds $70\%$. 

Fig. \ref{fig:full} shows generalization error and mutual information behaviour for experiments on MNIST and CIFAR-10 datasets. It is expected that both the mutual information and the generalization error have a decreasing behavior. Both decreasing behaviors were experimentally corroborated, but both curves have a similar shape in terms of decay. In addition, mutual information detects the slope changes of the generalization error, highlighting the link between these two magnitudes. 

\subsection{Mutual Information in a unstructured dataset}

In order to check that the conclusions extracted in the previous experiments were not exclusive to image dataset, in this Appendix we study the relation between generalization error and mutual information for an unstructured dataset.

The forest cover dataset \citep{arbolitosdataset} or \emph{covertype dataset} are cartographic data about types of forests in the Roosevelt National Forest in Colorado from the UCI KDD archive\footnote{\url{https://archive.ics.uci.edu/ml/datasets/covertype}}. This dataset includes information on tree type, shadow coverage, distance to nearby landmarks (roads etcetera), soil type, and local topography. It is an unstructured dataset of $54$ features and presentes a supervised classification task to classify each observation into one of seven mutually exclusive forest cover type classes\footnote{The seven forest cover type classes used in this study were lodgepole pine (Pinus contorta), spruce/fir (Picea engelmannii and Abies lasiocarpa), ponderosa pine (Pinus ponderosa), Douglas-fir (Pseudotsuga menziesii), aspen (Populus tremuloides), cottonwood/willow (Populus angustifolia, Populus deltoides, Salix bebbiana, Salix amygdaloides), and krummholz.}. We random split samples at $80\%$ training and $20\%$ testing and we normalized the input features in mean and variance.

We considered a Normal encoder (see Section \ref{sec:normal_encoder}) composed of $50$ hidden units with ReLU activation followed by another linear layer for each parameter ($\mu$ and $\log\sigma^2$) and with $35$ hidden units. That is each parameter, $\mu$ and $\log\sigma^2$, are a two-layers network where the first one is common to both. We chose a learning rate of $0.001$, a batch-size of $1024$ and trained during $200$ epochs. As covertype is a strongly unbalanced dataset, we oversampling data with SMOTE technique \citep{smote}. Due to the high variability of the dataset, we initialized the $\log\sigma^2$ layer with a truncated normal of $0.01$ standard deviation in order to prevent numerical explosions. We repeated simulations $4$ times for each Lagrange multiplier value and report the average metrics.

\begin{figure}[t]
\begin{subfigure}{.46\textwidth}
  \centering
  \includegraphics[width=\linewidth]{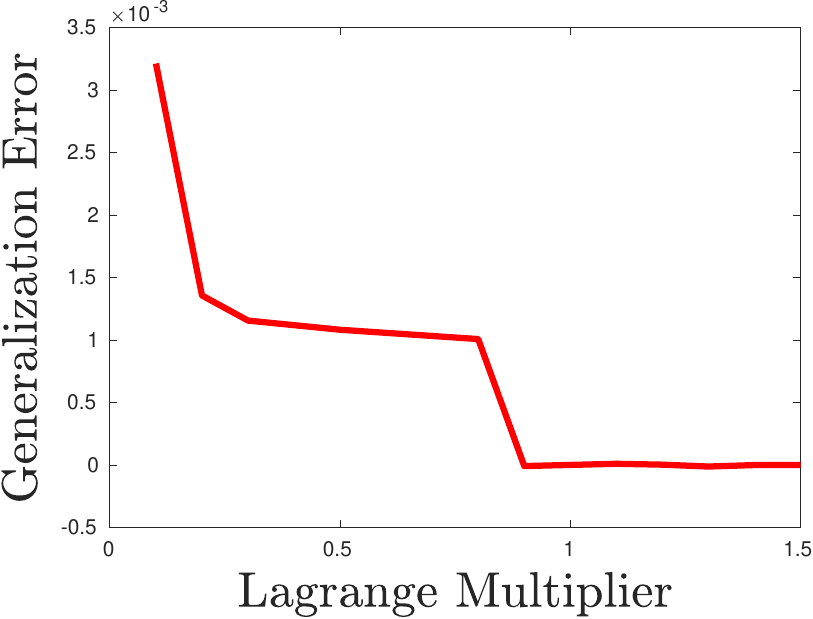}  
  \caption{Generalization Error}
\label{fig:forest_gap}
\end{subfigure}
\begin{subfigure}{.46\textwidth}
  \includegraphics[width=\linewidth]{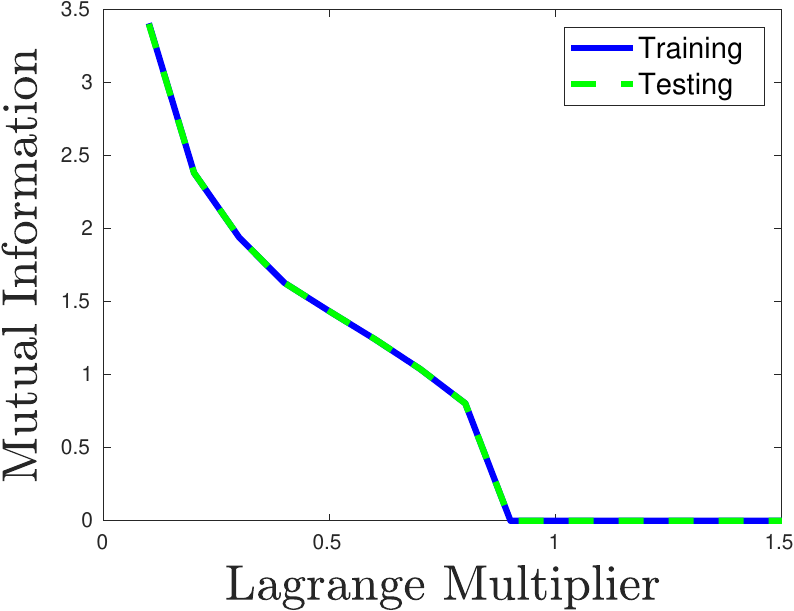}  
  \caption{Mutual Information}
\label{fig:forest_mi}
\end{subfigure}
\caption{Generalization error (a) and mutual information (b) for Forest Cover dataset.}
\label{fig:forest}
\end{figure}

Fig. \ref{fig:forest} shows generalization error and mutual information behaviour. Both magnitudes have a decreasing behaviour, as it is expected, but with a certain similarity. Curves have an strong slope change in $\lambda\in(0.2,0.3)$ and both abruptly decrease for $\lambda\geq 0.9$.
}
\section*{Funding and/or Competing interests}
This work was partially supported by the projects PIP 11220150100578CO and UBACyT 20020170100470BA. This project has received funding from the European Union’s Horizon 2020 research and innovation program under the Marie Skłodowska-Curie grant agreement No 792464. The authors have no competing interests to declare that are relevant to the content of this article.

All authors contributed to the study conception and design. Analysis was performed by all authors. Material preparation and data collection were performed by Matias Vera. The first draft of the manuscript was written by Matias Vera and all authors commented on previous versions of the manuscript. All authors read and approved the final manuscript.
\bibliographystyle{spbasic}      
\bibliography{mybibfile}

\end{document}